%% file: main.tex
\documentclass[10pt,twocolumn,letterpaper]{article}

\input{math_commands.tex}

\usepackage{iccv}
\usepackage{times}
\usepackage{epsfig}
\usepackage{graphicx}
\usepackage{amsmath}
\usepackage{amssymb}

\usepackage{url}
\usepackage[utf8]{inputenc} 
\usepackage[T1]{fontenc}    
\usepackage[table,xcdraw,dvipsnames]{xcolor}
\usepackage{subcaption}
\usepackage[normalem]{ulem} 
\usepackage[export]{adjustbox}
\usepackage{booktabs}       
\usepackage{amsfonts}       
\usepackage{nicefrac}       
\usepackage{microtype}      
\usepackage{gensymb}
\usepackage{tabularx}
\usepackage{siunitx}
\usepackage{enumitem}
\usepackage{lineno}
\usepackage{pifont}
\usepackage[square,numbers]{natbib}

\usepackage[dvipsnames]{xcolor}

\usepackage[font=small,labelfont=bf]{caption}
\usepackage{multirow}
\usepackage{lipsum}
\usepackage{xhfill}

\usepackage[pagebackref=true,breaklinks=true,letterpaper=true,colorlinks,citecolor=ForestGreen,bookmarks=false]{hyperref}

\setcitestyle{square}
\setcitestyle{citesep={,}}
\iccvfinalcopy 

\usepackage{fancyhdr}

\fancypagestyle{firstpage}{%
  \fancyhf{}
  \chead{\small In Proceedings of IEEE International Conference on Computer Vision (ICCV), 2021}
}

\begin{document}

\title{Move2Hear: Active Audio-Visual Source Separation}


\author{Sagnik Majumder$^{1}$ \hspace{5mm} Ziad Al-Halah$^{1}$ \hspace{5mm} Kristen Grauman$^{1,2}$\\
$^1$The University of Texas at Austin \hspace{3mm} $^2$Facebook AI Research
\\
{\tt\small \{sagnik, ziad, grauman\}@cs.utexas.edu}
}

\maketitle

\thispagestyle{firstpage}

\input{sections/abstract}

\input{sections/introduction}

\input{sections/related_work}

\input{sections/task}

\input{sections/approach}

\input{sections/experiment}

\input{sections/conclusion}

{\small
\bibliographystyle{ieee_fullname}
\bibliography{main}
}

\newpage
\clearpage
\input{sections/supp}

\end{document}

%% file: math_commands.tex
\usepackage{amsmath,amsfonts,bm}

\def\eqref#1{equation~\ref{#1}}

\def\1{\bm{1}}

\DeclareMathAlphabet{\mathsfit}{\encodingdefault}{\sfdefault}{m}{sl}
\SetMathAlphabet{\mathsfit}{bold}{\encodingdefault}{\sfdefault}{bx}{n}



%% file: sections/abstract.tex
\begin{abstract}

We introduce the active audio-visual source separation problem, where an agent must move intelligently in order to better isolate the sounds coming from 
an object of interest in its environment.  The agent hears multiple audio sources simultaneously (e.g., a person speaking down the hall in a noisy household) and it must use its eyes and ears to automatically separate out the sounds originating from a target object within a limited time budget.
Towards this goal, we introduce a reinforcement learning approach that trains movement policies controlling the agent's camera and microphone placement over time, guided by the improvement in predicted audio separation quality.  We demonstrate our approach in scenarios motivated by both augmented reality (system is already co-located with the target object) and mobile robotics (agent begins arbitrarily far from the target object).  Using state-of-the-art realistic audio-visual simulations in 3D environments, we demonstrate our model's ability to find minimal movement sequences with maximal payoff for audio source separation. Project: \url{http://vision.cs.utexas.edu/projects/move2hear}.
\end{abstract}

%% file: sections/introduction.tex
\section{Introduction}

Audio-visual events play an important role in our daily lives.  However, in real-world scenarios, physical factors can either restrict or facilitate our ability to perceive them. For example, a father working upstairs might move to the top of the staircase to better hear what his child is calling out to him from below; a traveler in a busy airport may shift closer to the gate agent to catch the flight delay announcements amidst the din, without moving too far in order to keep her suitcase in sight; a friend across the table in a noisy restaurant may tilt her head to hear the dinner conversation more clearly, or scooch her chair to better catch music from the band onstage.

Such examples show how \emph{controlled sensor movements} can be critical for audio-visual understanding. In terms of audio sensing, a person's nearness and orientation relative to a sound source affects the clarity with which it is heard, especially when there are other competing sounds in the environment.  
In terms of visual sensing, one must see obstacles to 
circumvent them, spot desired and distracting sound sources, use visual context to hypothesize 
an out-of-view sound source's location, and actively look for ``sweet spots" in the visible 3D scene that may permit better listening.

\begin{figure}[t] 
    \centering
    \includegraphics[width=.9\linewidth]{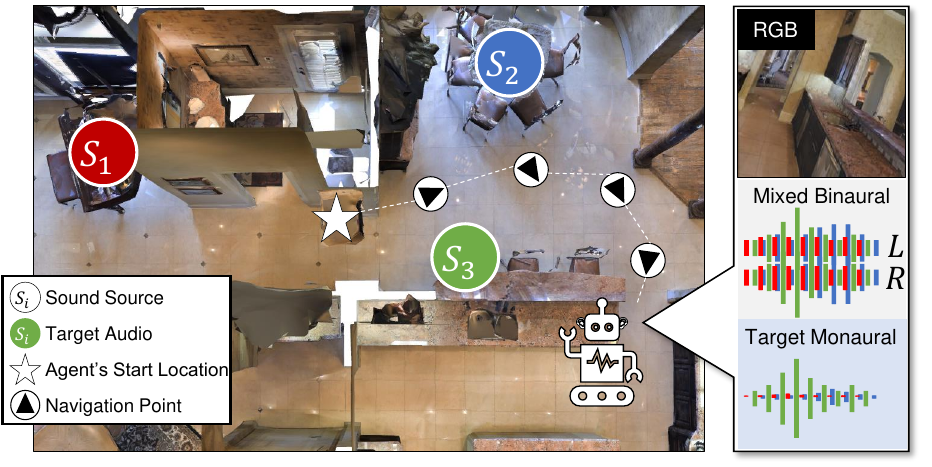}
    \vspace*{-0.1in}
    \caption{Active audio-visual source separation. Given multiple mixed audio sources $S_i$ in a 3D environment, the agent is tasked to separate a target source (shown in green) by intelligently moving around using cues from its egocentric audio-visual input to improve the quality of the predicted target audio signal. See text.
    }
\label{fig:intro}
\vspace{-0.5cm}
\end{figure}

In this work, we explore how autonomous multi-modal systems might learn to exhibit such intelligent behaviors.  In particular, we introduce the task of \emph{active audio-visual source separation}: given a stream of egocentric audio-visual observations, an agent must decide how to move in order to recover the sounds being emitted by some target object, and it must do so within bounded time. See Figure~\ref{fig:intro}. Unlike traditional audio-visual source separation, where the goal is to isolate sounds in passive, pre-recorded video~\cite{gabbay2017visual, gao2018learning, afouras2018conversation, ephrat2018looking, owens2018audio, gao2019co, chung2020facefilter, gao2021visualvoice}, the proposed task calls for actively controlling the camera and microphones' positions over time.  Unlike recent embodied AI work on audio-visual navigation, where the goal is to travel to the location of a sounding object~\cite{chen2020soundspaces,gan2019look,chen2020semantic,chen2021learning}, the proposed task calls for returning a separated soundtrack for an object of interest in limited time, without necessarily traveling all the way to it. 

We consider two variants of the new task. In the first, the system begins exactly at the location of the desired sounding object and must fine-tune its positioning to hear better; this variant is motivated by augmented reality (AR) applications where the object of interest is known and visible (e.g., the person seated across from someone wearing an assistive audio-visual AR device) yet local movements of the device sensors are still beneficial to improve the audio separation. In the second variant, the system begins at an arbitrary position away from the object of interest; this variant is motivated by mobile robotics applications, where an agent detects a sound from afar (e.g., the child calling from downstairs) but it is entangled with distractors and requires larger movements within the environment to hear correctly. We refer to these scenarios as \emph{near-target} and \emph{far-target}, respectively.

Towards addressing the active audio-visual source separation problem, we 
propose Move2Hear, a reinforcement learning (RL) framework 
where an agent learns a policy for how to move to 
hear better. The agent receives an egocentric audio-visual observation sequence (RGB and binaural audio) 
along with the target category of interest\footnote{e.g., a human speaker, a musical instrument, or some other sound type}, and at each time step decides its next motion (a rotation or translation of the camera and microphones).  During training, as it aggregates these observations over time 
with an explicit memory module and recurrent network, the agent is rewarded for improving its estimate of the target object's latent (monaural) sound. In particular, the reward promotes movements that better resolve the target sound from the other distractor sounds in the environment.  Our approach handles both 
near- and far-target scenarios.  

Importantly, optimal positioning for audio source separation is \emph{not} the same as navigation to the source, both because the agent faces a time budget---it may be impossible to reach the target in that time---and also because the geometry of the 3D environment and relative positions of distractor sounds make certain positions relative to the target more amenable to separation.
For example, in Fig.~\ref{fig:intro} the agent is tasked with separating the audio 
from object $S_3$.
Here, going directly to $S_3$ is not 
ideal, 
as that position will have high interference from the other audio sources in the scene, $S_1$ and $S_2$. By moving around the kitchen bar, the agent manages to dampen the signal from $S_1$ significantly due to the intermediate obstacles (walls) and, 
simultaneously, emphasize the signal from $S_3$ compared to $S_2$, 
thus leading to better separation. 

We 
test our approach 
with realistic audio-visual SoundSpaces~\cite{chen2020soundspaces} simulations on the Habitat platform~\cite{habitat19iccv} 
comprising 47 real-world 
Matterport3D 
environment scans~\cite{Matterport3D}, 
along with an array of sounds from diverse human speakers, music, and other background distractors. Our model successfully learns how to move to hear its target more clearly in unseen 
scenes, surpassing baselines that systematically survey their surroundings, 
smartly or randomly explore, or 
navigate directly to the source. We explore the synergy of both vision and audio for solving this task. 

Our main contributions are 1) we define the active audio-visual separation task, a new direction for embodied AI research; 2) we present the first approach to begin tackling this task, namely a new RL-based framework that integrates sound separation and visual navigation motion policies; and 3) we thoroughly experiment with a variety of sounds, visual environments, and use cases. While just the first step in this area, we believe our work lays groundwork to explore new problems for multi-modal agents that move to hear better.

%% file: sections/related_work.tex
\section{Related Work}\label{sec:related}

\paragraph{Passive Audio(-Visual) Source Separation.}
Passive (non-embodied) separation of audio sources using solely audio inputs has been  extensively studied in signal processing. While sometimes only single-channel monaural audio is assumed~\cite{10.1007/978-3-540-74494-8_52, Spiertz09source-filterbased, Virtanen07monauralsound, 6853860}, multi-channel audio captured with multiple microphones~\citep{nakadai2002real, Yilmaz04blindseparation, duong2010under}---including binaural audio~\citep{deleforge:hal-00768668, weiss2009source, zhang2017deep}
---facilitates  separation by making the spatial cues explicit. Using vision together with sound improves separation. Audio-visual (AV) separation methods leverage mutual information~\citep{hershey2000audio, NIPS2000_11f524c3}, subspace analysis~\citep{Smaragdis03audio/visualindependent, pu2017audio}, matrix factorization~\citep{7951787, 7760220, gao2018learning}, correlated onsets~\citep{4270342, 7952688}, and deep learning to separate speech~\citep{afouras2018conversation, gabbay2017visual, ephrat2018looking, owens2018audio, afouras2019my, chung2020facefilter, gao2021visualvoice}, music~\citep{gao2019co, gan2020music, xu2019recursive, zhao2018sound}, and other objects~\citep{gao2018learning}. While some methods extract the audio tracks for all classes present in the mixture~\citep{8683007, tzinis2020improving}, others isolate one specific target~\citep{DBLP:conf/interspeech/OchiaiDKIKA20, tzinis2020into, DBLP:conf/interspeech/GuCZZXYSZ019, gu2020temporal, 8736286}.

Whereas prior work assumes a pre-recorded video as input, our work addresses a new \emph{embodied perception} version of the audio separation task, in which an agent can see, hear, and move in a 3D environment to actively hear a source better.  To our knowledge, our work is the first to consider how intelligent movement influences a multi-modal mobile agent's ability to separate sound sources.  In addition, whereas existing video methods use
dynamic object motion to tease out audio-visual associations (especially for speech~\cite{gabbay2017visual, afouras2018conversation, ephrat2018looking, owens2018audio, chung2020facefilter, gao2021visualvoice}), our setting demands using visual cues in the surrounding 3D environment to move to ``sweet spots" for listening to a source amidst competing background sounds. Finally, our task requires recovering the target object's latent monaural audio as output---stripping away the effects of the agent's relative position, environment geometry, and scene materials. This aspect of the task is by definition absent for AV separation in passive video.

\vspace{-0.3cm}
\paragraph{Visual and Audio-Visual Navigation.} 
While mobile robots traditionally navigate by a mixture of explicit mapping and planning~\citep{thrun2002probabilistic, FuentesPacheco2012VisualSL}, recent work explores learning navigation policies from egocentric image observations (e.g.,~\citep{gupta2017cognitive, savinov2018semiparametric, mishkin2019benchmarking}).  Facilitated by fast rendering platforms~\cite{habitat19iccv} and realistic 3D visual assets~\cite{Matterport3D, xiazamirhe2018gibsonenv, replica19arxiv}, researchers develop reinforcement learning architectures to tackle 
diverse visual navigation tasks~\cite{gupta2017cognitive, mishkin2019benchmarking, wijmans2019dd, zhu-iccv2017, yang2019visual, batra2020objectnav, wu2019bayesian, mousavian2019visual, chaplot_object_2020, chang2020semantic,7989381, savinov2018semiparametric, neural-topo}. Going beyond purely visual agents, recent work explores joint audio-visual sensing for embodied AI~\citep{gan2019look, chen2020soundspaces, visual-echoes, purushwalkam2020audio, chen2020semantic, chen2021learning}. In the \emph{audio-visual navigation} task, an agent enters an unmapped environment and must travel to a sounding target object (e.g., go to the ringing phone)~\citep{gan2019look, chen2020soundspaces, chen2020semantic, chen2021learning}. Related efforts explore audio-visual spatial sensing to infer environment geometry~\cite{visual-echoes} or floorplans~\cite{purushwalkam2020audio}, or attempt audio-only navigation to multiple fixed-position sources in a gridworld while accounting for distractor sounds~\cite{Ranadive2020OtoWorldTL}.  

Our \emph{separation} goal is distinct from \emph{navigation}: our agent succeeds if it accurately separates the true target sound, not if it simply travels to where the sound or target object is. 
As we will show, the two tasks yield agents with differing behavior.  In fact, our model remains relevant even in the near-target scenario, where (unlike AV navigation) the position of the target is already known.  Compared to any of the above, our key novel insight is that audio-visual cues can tell an agent how to move to separate 
multiple active audio sources.

\vspace{-0.3cm}
\paragraph{Source Localization in Robotics.}

Robotic systems use microphone arrays to perform sound source localization, often via signal processing techniques on the audio stream alone~\cite{nakadai1999sound,rascon2017localization}.
To focus on a sound, such as a human speaker, the microphone can be actively steered towards the localized source   (e.g.,~\cite{asano2001,nakadai2000active,bustamante2018}).  Using both visual and audio cues to localize people and detect when they are speaking~\cite{alameda2015vision,viciana2014audio,ban2018icassp} is an important precursor to human-robot systems that follow conversations.
The proposed task also requires actively attending to audio events, but in our case there are multiple competing sound sources and they may be initially far from the agent. Our technical contribution is also complementary to existing methods: our approach learns to map audio-visual egocentric observations directly to long-term sequential actions. Learning behaviors from data, as opposed to fixing heuristics, offers potential advantages for generalization.

%% file: sections/task.tex
\section{Active Audio-Visual Source Separation}\label{sec:task}

We propose a novel task: active audio-visual source separation (AAViSS). In this task, an autonomous agent simultaneously hears multiple audio sources of different types (e.g., speech, music, background noise) that are 
at various locations in a 3D environment. The agent's goal is to separate a \emph{target source} (e.g., a specified human speaker or instrument) from the heard audio mixture by intelligently moving in the environment.
This \emph{active listening} task requires the agent to leverage both acoustic and visual cues. While the acoustic signal carries rich information about the types of audio sources and their relative distances and orientations from the agent, the visual signal is crucial both to see obstacles affecting navigation and identify useful locations from which to sample acoustic information in the visible 3D scene. 

\vspace{-0.3cm}
\paragraph{Task Definition.} 
In each episode of agent experience, multiple audio sources are randomly initialized in the 3D environment. The map of the environment is unknown to the agent as are the locations of the audio sources. At each step, the agent hears a mixed binaural
audio signal that is a function of the source types (e.g., human voice, music, etc.), their displacement from the agent, and their sound reflections resulting from the major geometric surfaces and materials in the 3D scene. One of the audio sources is the target, i.e., the source the agent wants to hear, as relevant to its overarching application setting. The agent is tasked with predicting the target's \emph{monaural} audio signal as clearly as possible---that is, the true latent target sound itself, separated from the other sources and independent of the spatial effects of where it is emitted. The agent must intelligently move and sample visual-acoustic cues from its environment to best predict the target signal by the end of a fixed time budget.

Note that it is significant that we define the correct output to be the \emph{monaural} target sound. Were the objective instead to output the \emph{binaural} sound of the target at the agent's current position, trivial but non-useful solutions would exist (e.g., moving to a position where the target is inaudible, and hence its binaural waveforms are approximately 0).

As discussed above, we consider two variants of this task depending on the agent's starting position relative to the target audio source. In the \emph{near-target} variant, the agent starts at the target and needs to conduct a sequence of fine-grained motions to extract the best target audio; in the \emph{far-target} variant, the agent starts in a random far position and must first navigate to the vicinity of the target before commencing movements for better separation.

\vspace{-0.3cm}
\paragraph{Episode Specification.} 
Formally, an episode is defined by a tuple $(E, p_0, S_1, S_2, \dots S_k, G^y)$ where $E$ is a 3D scene, $p_0=(l_0, r_0)$ is the initial agent pose defined by its location $l$ and rotation $r$, $S_i=(S_i^w, S_i^l, S_i^y)$ is an audio source defined by its periodic monaural waveform $S_i^w$, its location $S_i^l$, and type $S_i^y$. There are $k$ audio sources in the scene, each from a different type\footnote{Note that distinct human voices count as distinct types.} ($S_1^y\neq S_2^y \neq \dots \neq S_k^y$), and $G^y$ is the target audio goal type such that $G\in\{S_i\}$. At each step, the agent hears a mixture of all audio sources, and the goal is to predict $G^w$ by the end of the episode,  given the target goal label $G^y$. 
The episode length is $\mathcal{T}$ steps, meaning the agent has bounded time to provide its output.

\vspace{-0.3cm}
\paragraph{Action Space.}
The agent's action space $\mathcal{A}$ consists of $MoveForward$,  $TurnLeft$, and  $TurnRight$. At each step, the agent samples an action $a_t\in\mathcal{A}$ to move on a \emph{navigability graph} of the environment; that graph is unknown to the agent. While turning actions are always valid, $MoveForward$ is allowed only if there is an edge connecting the current node and the next one, and the agent is facing the destination node. Non-navigable connections exist due to walls and obstacles, e.g., a sofa blocking the path.

\vspace{-0.3cm}
\paragraph{3D Environment and Audio-Visual Simulator.}
Consistent with substantial embodied AI research in the computer vision community (e.g.,~\citep{chen2020soundspaces,ramakrishnan2020occant,gupta2017cognitive,savinov2018semiparametric}), and in order to provide reproducible results, we develop our approach using state-of-the-art visually and acoustically realistic 3D simulators. We use the SoundSpaces~\citep{chen2020soundspaces}  audio simulations, built on top of the AI-Habitat simulator~\citep{habitat19iccv} and the Matterport3D scenes~\citep{Matterport3D}. The Matterport3D scenes are real-world homes and other indoor environments with both 3D meshes and image scans. SoundSpaces provides room impulse responses (RIR) at a spatial resolution of 1 meter for the Matterport3D scenes. These state-of-the-art RIRs capture how sound from each source propagates and interacts with the surrounding geometry and materials, modeling all of the major real-world features of the RIR: direct sounds, early specular/diffuse reflections, reverberations, binaural spatialization, and frequency dependent effects from materials and air absorption. Our experiments further push the realism by considering noisy sensors (Sec.~\ref{sec:experiments}). See the Supp. video to gauge realism.

We place $k$ monaural audio sources in the 3D environment. Since current simulators do not support dynamic object rendering (e.g., people talking, instruments being played), these sources are represented as point objects~\citep{chen2020soundspaces}.

At each time step, we simulate the binaural mixture of the $k$ sounds coming from their respective locations in the scene, as received by the agent at its current position. Specifically, the sources' waveforms $S_i^w$ are convolved with RIRs corresponding to the scene $E$ and the agent pose and the source location pairs $(p, S_i^l)$. Subsequently, the output of the convolved RIRs is mixed together to generate dynamic binaural mixtures as the agent moves around:
\vspace*{-0.09in}
\begin{equation}
    B^{w, mix}_{t} = \frac{1}{k}\sum_{i=1}^{k}{B^{w, i}_{t}},
\vspace*{-0.09in}
\end{equation}
where $B^{w, i}_{t}$ is the binaural waveform of the sound source $i$ at time $t$, and $B^{w, mix}_{t}$ is the binaural waveform of the mixed audio. Note that the ground truth  
waveforms per source are known only by the simulator; during inference, the agent observes only the mixed binaural sound, which changes with $t$ as a function of the agent's movements.

%% file: sections/approach.tex
\section{Move2Hear Approach}
\label{approach}
\begin{figure}[t] 
    \centering
    \includegraphics[width=1.\linewidth]{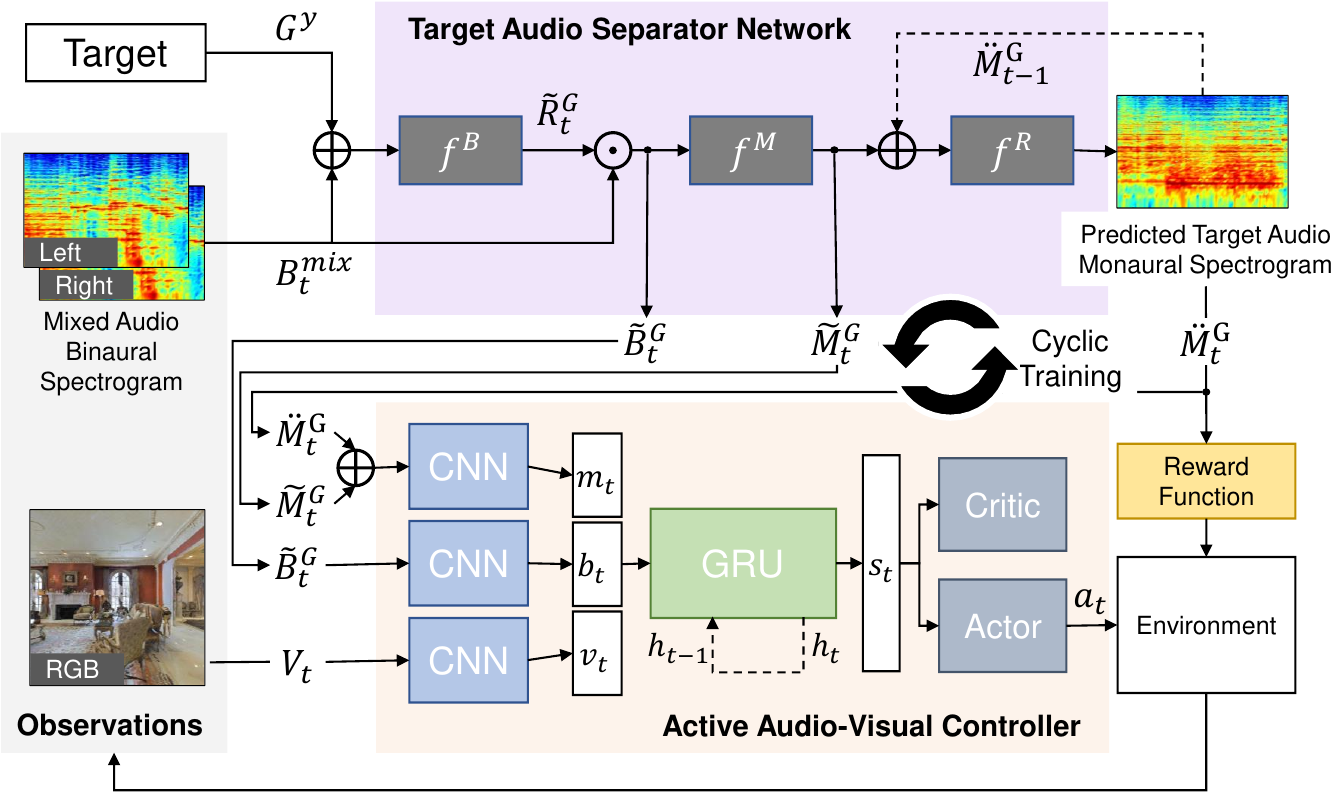}
    \caption{Our model for active audio-visual source separation has two main components: 1) an audio separator network (top) and 2) an active audio-visual controller (bottom). At each step, our model receives mixed audio from multiple sources in the 3D environment along with egocentric RGB views. The model actively moves in the environment to improve its separation of a target audio source.
    } 
    \label{fig:model}
    \vspace{-0.25cm}
\end{figure}

We pose the AAViSS task as a reinforcement learning problem, where the agent learns a policy to sequentially decide how to move given its stream of egocentric audio-visual observations.
Our model has two main components (see Fig.~\ref{fig:model}): 1) the target audio separator network and 2) the active audio-visual (AV) controller. 
The separator network has two functions: it separates the target audio signal from the heard mixture at each step, and it informs the controller about its current estimate to improve the separation. 
The AV controller learns a policy guided by the separation quality, such that it moves the agent in the 3D environment to improve the predicted target audio signal.  
These two components learn from each other during training to help the agent build an implicit understanding of how the separation quality changes given the current surrounding
3D structure (furniture, walls, rooms, etc.), the various audio sources, and their (unobserved) locations relative to the agent and the target. Hence, they enable the agent to learn a useful active movement policy to improve audio separation quality.

\subsection{Target Audio Separator Network}
\label{approach:audio_net}

At each step $t$, the audio network $f^A$ receives the mixed binaural sound $B^{mix}_t$ coming from all audio sources in the scene and the target audio type $G^y$, and it predicts the monaural target audio $\ddot{M}^G$, i.e., $f^A(B^{mix}_t, G^y)=\ddot{M}^G$ (Fig.~\ref{fig:model} top).   
We use the short-time Fourier transform (STFT)
to represent both the monaural $M$ and binaural $B$ audio spectrograms.
$M$ and $B$ are matrices, with $B \in \mathbb{R}_{+}^{2 \times F \times N}$ and $M \in \mathbb{R}_{+}^{F \times N}$ where $F$ is the number of frequency bins, $N$ is the time window, and $B$ has two channels (left and right). 

The audio network $f^A$ predicts $\ddot{M}^G$ in three steps and using three modules, such that: $f^A = f^B \circ f^M \circ f^R$.
First, given the target category $G^y$, the binaural target audio separator $f^B$ separates the target's binaural signal $\tilde{B}^G_t$ from the input mixture $B^{mix}_t$.
Second, the monaural audio predictor $f^M$ takes the previous binaural output $\tilde{B}^G_t$ and predicts the monaural target audio $\tilde{M}^G_t$ (i.e., independent of the room acoustics and spatial effects). 
Finally, given the monaural estimates from  all previous steps  and the current one $\tilde{M}^G_t$, the acoustic memory refiner $f^R$ continuously enhances the target monaural audio prediction $\ddot{M}^G_t$.
Next we describe the architecture of these three modules in detail.

\vspace{-0.3cm}
\paragraph{Binaural Audio Separator.}

For the binaural extractor $f^B$, we use a multi-layer convolutional network in a U-Net like architecture~\citep{RFB15a} with ReLU activations in the last layer.
Specifically, we concatenate the target label $G^y$ with the mixed binaural audio $B^{mix}$ along the channel dimensions, and pass this input to the U-Net to predict a real-valued ratio mask $\tilde{R}^T$~\citep{gao2019co, xu2019recursive, zhao2018sound} for the target binaural spectrogram:
\begin{equation}\label{eq:f_B}
    \tilde{R}^G_t = f^B(B^{mix}_{t} \oplus G^y),
\end{equation}
where $\oplus$ denotes channel-wise concatenation. 
Then, the predicted spectrogram for the target binaural $\tilde{B}^G_t$ is obtained by soft-masking the mixed binaural spectrogram with $\tilde{R}^G_t$: 
\begin{equation}
    \tilde{B}^G_t = \tilde{R}^G_t \odot B^{mix}_{t},
\end{equation}
where $\odot$ denotes the element-wise product of two tensors.

\vspace{-0.3cm}
\paragraph{Monaural Audio Predictor.}

Similarly, we use another U-Net for $f^M$ to predict the target monaural audio at the current step given the prediction from $f^B$, i.e.: 
\begin{equation}
    \tilde{M}^G_t = f^M(\tilde{B}^G_t).
\end{equation} 
$\tilde{M}^G_t$ serves as an initial estimate of the target monaural that our model predicts based only on the mixed binaural audio heard at the current step $t$. 
The factorization of monaural prediction into two steps $f^B$ and $f^M$ allows our model to first focus on the audio source separation by extracting the target audio in the same domain as the input (binaural) using $f^B$ then to learn how to remove spatial effects from the binaural to get the monaural signal using $f^M$.
Additionally, $f^B$ provides the policy with spatial cues about the target to help the agent anchor its actions in relation to the target location (see Sec.~\ref{approach:policy}). 
We refine this prediction using an acoustic memory, as we will describe next.

\vspace{-0.3cm}
\paragraph{Acoustic Memory Refiner.}

The acoustic memory refiner $f^R$ is a CNN that receives the current monaural separation $\tilde{M}^G_t$ from $f^M$ and its own previous prediction $\ddot{M}^G_{t-1}$ as inputs, and predicts the refined monaural audio $\ddot{M}^G_t$:
\begin{equation}
    \ddot{M}^G_t = f^R(\tilde{M}^G_t \oplus \ddot{M}^G_{t-1}).
\end{equation}
See Fig.~\ref{fig:model} top right.
The acoustic memory plays an important role in stabilizing the monaural predictions and helping the agent learn a useful policy, and it also provides robustness to microphone noise.
By taking into consideration the previous estimate $\ddot{M}^G_{t-1}$ and its relation to $\tilde{M}^G_t$, the model can learn when to update the monaural estimate of the target and by how much.  
This encourages non-myopic behavior: it allows the agent to explore its vicinity with less pressure to 
improve the quality of the predictions, in case there is a need to traverse intermediary low quality spots in the environment. 
Consequently, when the navigation policy is trained with a reward that is a function of the improvement in $f^R$'s prediction quality (details below), the agent learns to visit spaces in the scene that will improve the separation quality over time.

All three modules $f^B$, $f^M$, and $f^R$ are trained in a supervised manner using the ground truth of the target binaural and monaural signal, as detailed in Sec.~\ref{approach:training}.

\subsection{Active Audio-Visual Controller}
\label{approach:policy}

The second component of our approach is an AV controller  that  guides the agent in the 3D environment to improve its audio predictions (Fig.~\ref{fig:model} bottom).
The controller leverages both visual and acoustic cues to predict a sequence of actions $a_t$ that will improve the output of  $f^A$, the separator network defined above.
It has two main modules: 1) an observation encoder and 2) a policy network.

\vspace{-0.3cm}
\paragraph{Observation Space and Encoding.}
At every time step $t$, the AV controller receives the egocentric RGB image $V_{t}$, the current binaural separation $\tilde{B}^G_t$ from $f^B$, and the channel-wise concatenation of the target monaural predictions from $f^M$ and $f^R$, that is, $\bar{M}^G_t = \tilde{M}^G_t \oplus \ddot{M}^G_t$.

The audio and visual inputs carry complementary cues required for efficient navigation to improve the separation quality. 
$\tilde{B}^G_t$ conveys spatial cues about the target (its general direction and distance relative to the agent) which helps the agent to anchor its actions in relation to the target location.  Importantly, the better the separation quality of $\tilde{B}^G_t$, the more apparent this directional signal is (compared to $B_t^{mix}$, the mixed audio directly observed by the agent).
$\bar{M}^G_t$ is particularly useful in letting the policy learn the associations between the model's current position and the quality of the prediction in that position (captured by $\tilde{M}^G$), the overall quality so far (captured by $\ddot{M}^G$), and whether this position led to a change in the estimate (captured by $\bar{M}^G$). 

The visual signal $V_t$ provides the policy with cues about the geometric layout of the 3D scene so that the agent can avoid colliding with obstacles. 
Further, the visual input coupled with the audio allows the agent to capture relations between the 3D scene and the expected separation quality at different locations.   

We encode the three types of input using separate CNN encoders: $v_{t} = E^{V}(V_{t})$, $b_{t} = E^B(\tilde{B}^T_t)$ and $m_{t} = E^{M}(\bar{M}^G_t)$.
We concatenate the three feature outputs to obtain the current audio-visual encoding $o_t = [v_t, b_t, m_t]$.

\vspace{-0.3cm}
\paragraph{Policy Network.}
The policy network is made up of a gated recurrent network (GRU) that receives the current audio-visual encoding $o_t$ and the accumulated history of states $h_{t-1}$ to update its history to $h_{t}$ and outputs the current state representation $s_{t}$.  
This is followed by an actor-critic module that takes $s_{t}$ and $h_{t-1}$ as inputs to predict the policy distribution $\pi_{\theta}(a_{t} | s_{t}, h_{t-1})$ and the value of the state $\mathcal{V}_{\theta}(s_{t}, h_{t-1})$, where $\theta$ are the policy parameters. 
The agent samples an action $a_{t} \in \mathcal{A}$ according to $\pi_{\theta}$ to interact with its environment.

\vspace{-0.3cm}
\paragraph{Near- and Far-Target Policies.}
For the near-target task, we learn a \emph{quality policy} $\pi^{\mathcal{Q}}$ that is driven by the improvement in the target audio prediction (reward defined below). For the far-target variant, we learn a composite policy made up of $\pi^{\mathcal{Q}}$ and
an \emph{audio-visual navigation policy} $\pi^{\mathcal{N}}$ trained to get closer to the target audio.

The composite policy uses a time-based strategy to switch control between the two policies.
First, the navigation policy  
brings the agent closer to the target audio  
in a budget of $\mathcal{T}^{\mathcal{N}}$ steps,  
then the agent switches control to the quality policy to focus on improving the target audio separation. 
We found alternative blending strategies, e.g., switching based on predicted distance to the target, 
inferior in practice.
The audio network $f^A$ is active throughout the episode in both the near- and far-target tasks.

\subsection{Training}
\label{approach:training}

\paragraph{Training the Target Audio Separator Network.}
The separator network has two outputs: the binaural $\tilde{B}$ and the monaural audio predictions $\tilde{M}$ and $\ddot{M}$. 
We train it using the respective ground truth spectrograms of the target audio which are provided by the simulator: 
\begin{equation}\label{eq:loss_binaural}
    \mathcal{L}^B = ||\tilde{B}^G_t - B^G_t||_{1},
\end{equation}
where $B^G_t$ is the ground truth binaural spectrogram of the target at step $t$. 
Similarly, for the monaural predictions:
\begin{align}\label{eq:loss_monaural}
    \mathcal{L}^M &= ||\tilde{M}^G_t - M^G||_{1}, ~~~~~\mathcal{L}^R &= ||\ddot{M}^G_t - M^G||_{1},
\end{align}
where $M^G$ is the ground-truth monaural spectrogram for the target. Note that the predictions from $f^B$ and $f^M$ (i.e., $\tilde{B}^G_t$ and $\tilde{M}^G_t$ respectively) are step-wise predictions, unlike $f^R$ that takes into consideration the history of monaural predictions in the episode to refine its estimate.
Hence, we pretrain $f^B$ and $f^M$ using $\mathcal{L}^B$ and $\mathcal{L}^M$ and a dataset we collect from the training scenes.
For each datapoint in this dataset, we place the agent and $k$ audio sources randomly in the scene, then at the agent location we record the ground truth spectrograms ($B^G$, $M^G$) for a randomly sampled target type. We find this pretraining stage leads to higher performance and brings more stability to the predictions of the audio separator network compared to training those modules on-policy.

Once $f^B$ and $f^M$ are trained, we freeze their parameters and train $f^R$ on-policy along with the audio-visual controller, since the sequence of actions taken by the agent impact the history of the monaural predictions observed by $f^R$.

\vspace{-0.3cm}
\paragraph{Training the Active Audio-Visual Controller.}

The policy guides the agent to improve its audio separation quality by moving around.  Towards this goal, we formulate a novel dense RL reward to train the quality policy $\pi^\mathcal{Q}$:
\begin{equation}
    r_{t} = 
    \begin{cases}
        r^{s}_{t} &\quad\quad 1 \leq t \leq \mathcal{T} - 2\\
        -10 \times \mathcal{L}^R_\mathcal{T} + r^{s}_{t}  &\quad\quad t = \mathcal{T} - 1,\\
    \end{cases}
\end{equation}
where $r^{s}_{t} = \mathcal{L}^R_t - \mathcal{L}^R_{t + 1}$ is the step-wise reward that captures the improvement in separation quality of the monaural audio, and $r_{\mathcal{T}-1}$ 
is a one-time sparse reward at the end of the episode.
While $r^s_t$ encourages the agent to improve the separation quality at each step, the final reward $r_{\mathcal{T}-1}$ encourages the agent to take a trajectory that leads to an overall high-quality separation in the end. For the navigation policy $\pi^{\mathcal{N}}$, we adopt a typical navigation reward~\citep{habitat19iccv, chen2020soundspaces, chen2021learning} and reward the agent with +1.0 for reducing the geodesic distance to the target source and an equivalent penalty for increasing it. 

We train $\pi^\mathcal{Q}$ and $\pi^\mathcal{N}$ using Proximal Policy Optimization (PPO)~\citep{schulman2017proximal} with trajectory rollouts of 20 steps.  
The PPO loss consists of a value network loss, policy network loss, and entropy loss to encourage exploration (see Supp.). Both policies have the same architecture but differ in their reward functions and distribution of their initial agent locations $p_{0}$.

\vspace{-0.3cm}
\paragraph{Cyclic Training.}
We train the audio memory refiner $f^R$ and the policies $\pi^\mathcal{Q}$ and $\pi^\mathcal{N}$ jointly. We adopt a cyclic training scheme, i.e., in each cycle, we alternate between training the audio memory refiner and the policy for $U=6$ parameter updates. The cyclic training helps with stabilizing the RL training by ensuring partial stationarity of the rewards, particularly while training $\pi^{\mathcal{Q}}$, where the reward is a function of the quality of the target monaural predictions from $f^R$.

%% file: sections/experiment.tex
\section{Experiments}
\label{sec:experiments}

\paragraph{Experimental Setup.}
For each episode, we place $k=2$ (we also test with $k=3$) audio sources randomly in the scene at least 8 m apart, and designate one as the target.
The agent starts at the target audio location for the \emph{near-target} task, and at a random location 4 to 12 m from other sources for the \emph{far-target} task. The 12 m upper limit ensures that the agent can hear the target audio at the onset of its trajectory. We set the maximum episode length to $\mathcal{T}=20$ and $100$ steps for \emph{near-target} and \emph{far-target}, respectively. $\mathcal{T}^\mathcal{N}=80$ is set using the validation split. We use all 47 Matterport3D scenes that are large enough to generate at least 500 distinct episodes given the setup  above. We form train/val/test splits of 24/8/15 scenes and 112K/100/1K episodes. Because the test and train/val environments are disjoint, the agent is always tested in an unmapped space.

We use 12 types of sounds from three main groups: speech, music, and background sounds.
For speech, we sample 10 distinct speakers from the VoxCeleb1 dataset~\citep{Nagrani17} with different genders, accents, and languages. For music, we use a variety of instruments from the MUSIC dataset~\citep{zhao2018sound}. For background sounds, we sample non-speech and non-music sounds (e.g., clock-alarm, dog barking, washing machine) from ESC-50~\citep{piczak2015dataset}. The target in each episode can be one of $G\in\{Speakers, Music\}$, and the distractor(s) can be one of $D\in\{Speakers, Music, Background\}$ such that $G\neq D$ in the episode. Note that $Speakers$ 
denotes 10 separate speaker classes, resulting in a total of 11 target and 12 distractor classes. This enables us to evaluate a variety of audio separation scenarios: fine-grained separation (among the different speakers), coarse-grained separation (speech vs. music), and separation against background and ambient sounds commonly encountered in daily life. In total, we sample 23,677 1 sec audio clips of all types for use as monaural sounds. When testing on unheard sounds, we split the monaural sounds in the train:val:test with ratio 16:1:2. The longer audio clips used to produce the 1 sec unheard audio clips have no overlap among train, val, and test.

See Supp.~for all other details like spectrograms, network architectures, training hyperparameters, and baseline details.

\vspace{-0.4cm}
\paragraph{Baselines.} Since no prior work addresses the proposed task, we design strong baselines representing policies from related tasks and passive/un-intelligent motion policies:

\begin{itemize}[leftmargin=*,topsep=0pt,partopsep=0pt,itemsep=0pt,parsep=0pt]
\item\textbf{Stand In-Place:} audio-only baseline where the agent holds its starting pose for all steps, representing a default passive source separation method.
\item\textbf{Rotate In-Place:} audio-only baseline where the agent stays at the starting location and keeps rotating in place, i.e., sampling acoustic cues from different orientations.
\item\textbf{DoA:} Inspired by~\citep{nakadai2000active}, this agent faces the audio direction of arrival (DoA), i.e., it directs its microphones at the target sound from one step away (only relevant for \emph{near-target}). 
\item\textbf{Random:} an agent that randomly selects an action from the action space $\mathcal{A}$.
\item\textbf{Proximity Prior:} an agent that selects random actions but stays within a radius of 2 m (selected via validation) of the target so it cannot wander far from locations that are likely better for separation. 
Note that this baseline assumes an oracle for distance to target, not given to our method.
\item\textbf{Novelty~\citep{bellemare2016unifying}:} standard visual exploration agent trained to visit as many novel locations as possible within $\mathcal{T}$. 
\item\textbf{Audio-visual (AV) Navigator~\citep{chen2020soundspaces}}: state-of-the-art deep RL AudioGoal navigation agent~\citep{chen2020soundspaces} adapted for our task to additionally take the target audio category as input. Its audio input space exactly matches that of $f^B$ and it is trained with the typical navigation reward~\citep{habitat19iccv, chen2020soundspaces, chen2021learning}. 
\end{itemize}

For fair comparison, all baselines use our audio separator network $f^A$ as the audio separation backbone, taking as input the audio/visual observations resulting from their chosen movements in the scene. Specifically, all agents share the same $f^B$ and $f^M$, and only the audio memory refiner $f^R$ is trained online with its respective policy. This means that any differences in performance are attributable to the quality of each method's action selection.

\begin{table}
           \centering
            \begin{minipage}{.5\textwidth}
            \centering
              \scalebox{0.85}{
              \setlength{\tabcolsep}{2pt}
                \begin{tabular}{l cc|cc}
                \toprule
                &   \multicolumn{2}{ c| }{\textit{Heard}} &  \multicolumn{2}{ c }{\textit{Unheard}}\\
                Model  & {SI-SDR $\uparrow$} & {STFT $\downarrow$} & {SI-SDR $\uparrow$} & {STFT $\downarrow$}\\
                \midrule
                Stand In-Place      & 3.49  & 0.287 & 2.40  & 0.325 \\
                Rotate In-Place     & 3.45  & 0.285 & 2.50  & 0.321  \\
                DoA                 & 3.63  & 0.280 & 2.59  & 0.316 \\
                Random              & 3.68  & 0.280 & 2.57  & 0.319 \\
                Proximity Prior     & 3.74  & 0.276 & 2.63  & 0.315\\ 
                Novelty~\citep{bellemare2016unifying} & 3.82  & 0.276 & 2.86  & 0.318 \\  
                Move2Hear (Ours) & \textbf{4.31} & \textbf{0.260} & \textbf{3.20} & \textbf{0.298}\\ 
                \bottomrule
              \end{tabular}
              }
              \vspace{-0.1in}
              \caption{Near-Target AAViSS$^3$.  }\label{sub_table:near_target_2source_main}
            \end{minipage}
            
            \begin{minipage}{.5\textwidth}
            \vspace{0.05in}
            \centering
              \scalebox{0.85}{
              \setlength{\tabcolsep}{2pt}
                \begin{tabular}{l cc|cc}
                \toprule
                &   \multicolumn{2}{ c| }{\textit{Heard}} &  \multicolumn{2}{ c }{\textit{Unheard}}\\
                Model  & {SI-SDR $\uparrow$} & {STFT $\downarrow$} & {SI-SDR $\uparrow$} & {STFT $\downarrow$}\\
                \midrule
                    Stand In-Place & 0.74 & 0.390 & 0.09 & 0.416\\
                    Rotate In-Place           & 1.01 & 0.382 & 0.26 & 0.412\\
                    Random          & 1.15      & 0.378  & 0.46  & 0.402  \\
                    Novelty~\citep{bellemare2016unifying}   & 1.74      & 0.356  & 1.31  & 0.367 \\
                    AV Navigator~\citep{chen2020soundspaces}      & 1.46      & 0.368  & 0.72  & 0.396 \\
                    Move2Hear (Ours)       & \textbf{3.50} & \textbf{0.291} & \textbf{2.33} & \textbf{0.333} \\
                    \bottomrule
              \end{tabular}
              }
              \vspace{-0.1in}
              \caption{Far-Target AAViSS. }\label{sub_table:far_target_2source_main}
            \end{minipage}
            \vspace{-0.25in}
\end{table}

\vspace{-0.4cm}
\paragraph{Evaluation.}
We evaluate the target monaural separation quality at the end of $\mathcal{T}$ steps, for 1000 test episodes with 3 random seeds. We use the ground-truth monaural phase~\citep{simpson2015deep} for all methods and the inverse short-time Fourier transform to reconstruct a time-discrete monaural waveform from $\ddot{M}^G$.  We use standard metrics: \textbf{STFT} distance, a spectrogram-level measure of prediction error, and \textbf{SI-SDR}~\citep{8683855}, a scale-invariant measure of distortion in the reconstructed signal.

\subsection{Active Audio-Visual Source Separation}

\vspace{-0.1cm}
\paragraph{Near-Target.}
Table \ref{sub_table:near_target_2source_main} reports the separation quality of all models on the \emph{near-target} task.\footnote{AV Navigator~\cite{chen2020soundspaces} is not applicable here; the agent begins at the target.} Passive models that stay at the target (Stand, Rotate) do not perform as well as those that move (e.g., Proximity Prior). DoA fares better than the In-Place baselines as it gets to direct its microphones towards the target to sample a cleaner signal. Novelty outperforms the other baselines, showing the benefit of adding vision and sampling diverse acoustic cues. Our Move2Hear model outperforms all baselines by a statistically significant margin (according to the Kolmogorov–Smirnov test with $p\leq 0.05$). Move2Hear learns to take deliberate sequences of actions to improve the separation quality by reasoning about the 3D environment and inferred source locations. 

\begin{figure}[t]
    \centering
    \begin{subfigure}[b]{0.49\linewidth}
    \centering
    \includegraphics[width=\linewidth]{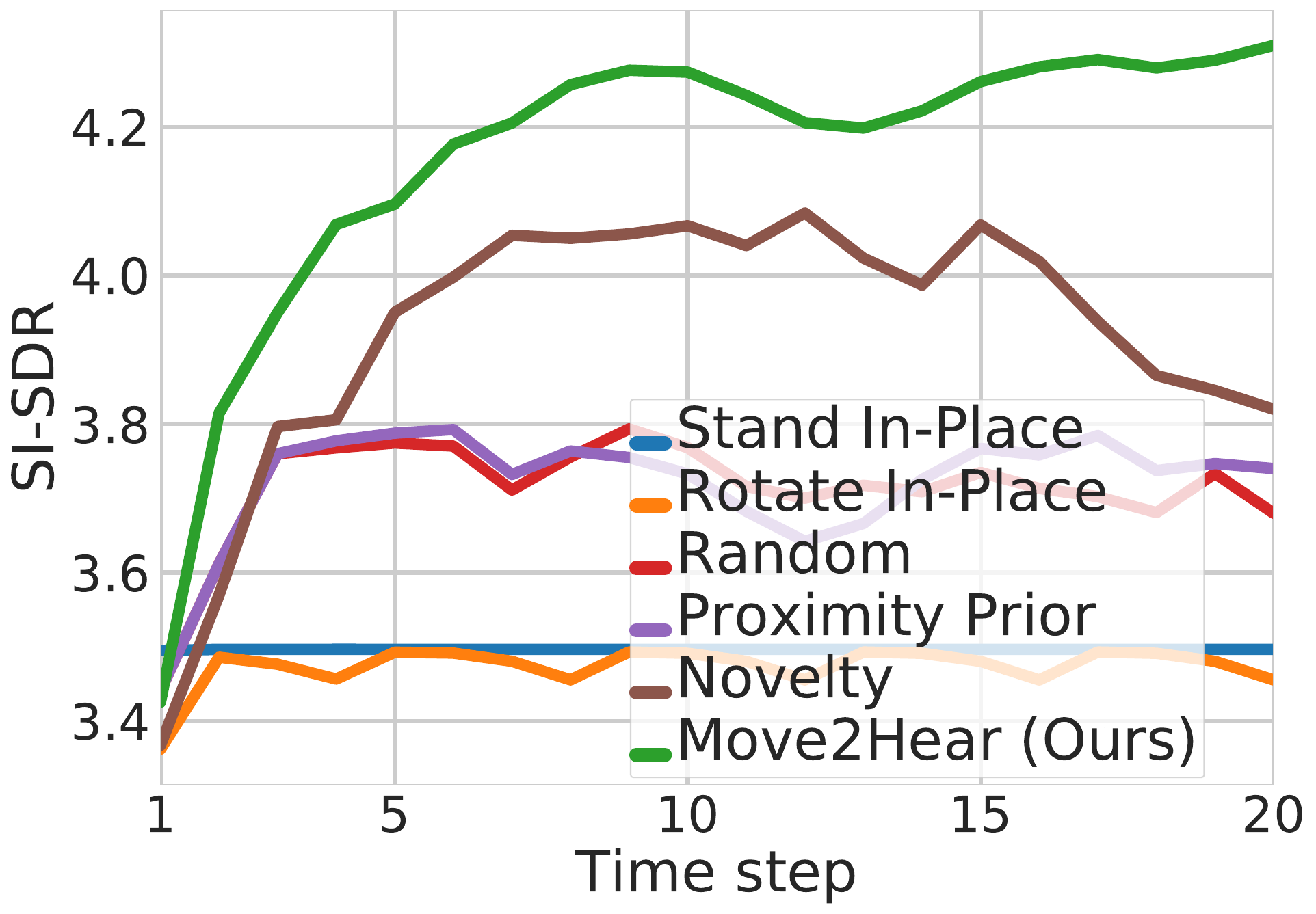}
    \caption{Separation progress}
    \label{fig:performance_vs_time_near_target_2_source}
    \end{subfigure}\hfill
    \begin{subfigure}[b]{0.49\linewidth}
    \centering
    \includegraphics[width=\linewidth]{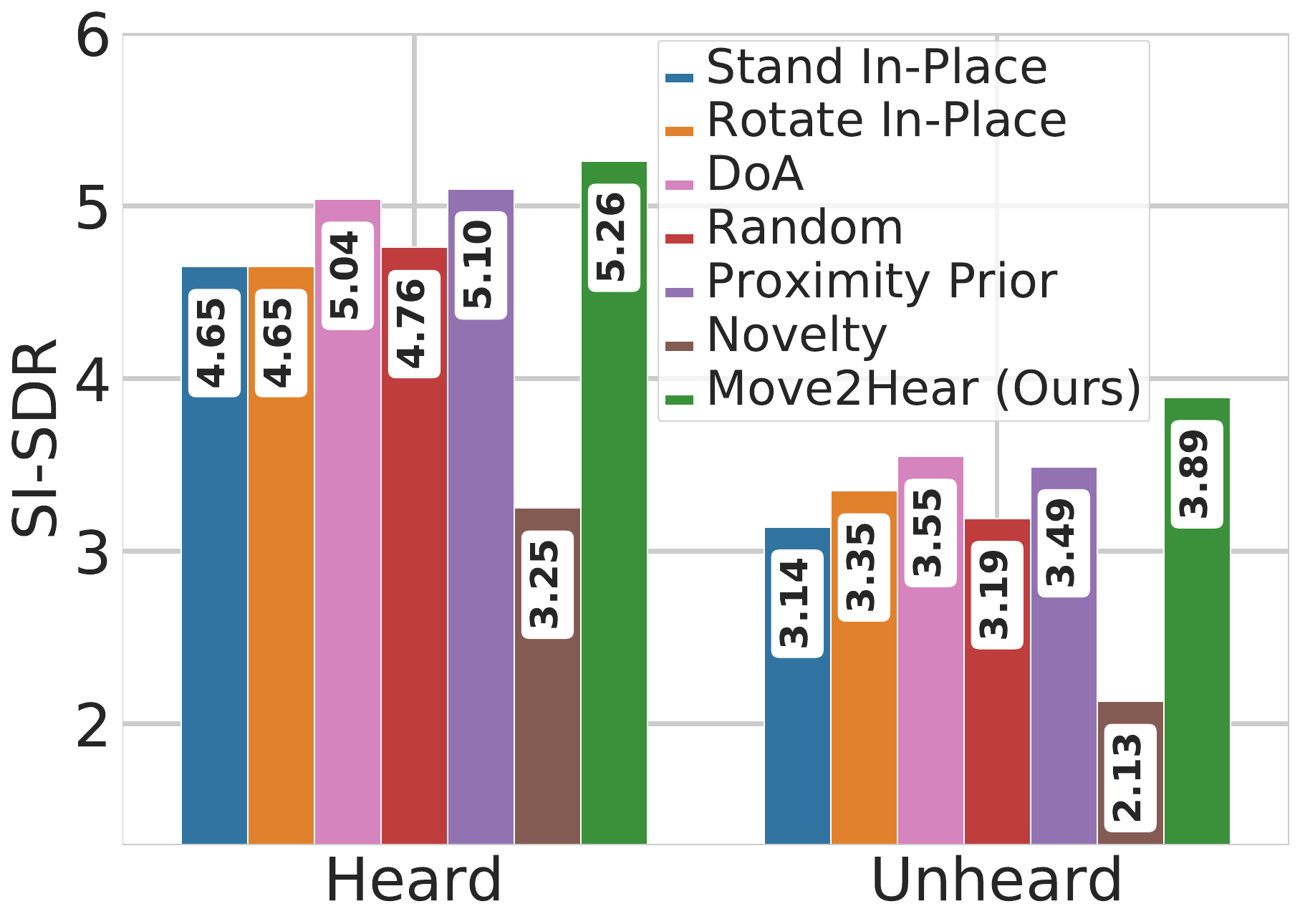}
    \caption{Separation with 3 sources}
    \label{fig:performance_vs_model_target_3_source}
    \end{subfigure}
    \vspace*{-0.3cm}
\caption{(a) Separation quality as a function of time. (b) Final separation performance with 3 sources (i.e., 2 distractors).}
    \label{fig:steps_k3}
    \vspace{-0.3cm}
\end{figure}

Fig.~\ref{fig:performance_vs_time_near_target_2_source} shows performance across each step in the episode. 
The non-stationary models make progress initially when sampling the cues close to the target, but then flatten quickly. In contrast, Move2Hear keeps improving with almost each action it takes, anticipating locations better for separation and learning behavior distinct from the other motion policies.

\vspace{-0.41cm}
\paragraph{Far-Target.}
Table~\ref{sub_table:far_target_2source_main} shows the results on the \emph{far-target} task.
Here again we see a distinct advantage for models that move around. Interestingly, AV Navigator~\citep{chen2020soundspaces} performs worse than Novelty~\citep{bellemare2016unifying} even though it has been trained to navigate towards the target. 
This highlights the difficulty of audio goal navigation in the presence of distractor sounds and the need for high-quality separations for successful navigation. Our model outperforms the previous baselines by a significant margin ($p \leq 0.05$).


\begin{table}[t]
  \centering
    \scalebox{0.85}{
    \setlength{\tabcolsep}{2pt}
    \begin{tabular}{lcc|cc}
    \toprule
                    &   \multicolumn{2}{c|}{\textit{Heard}} &  \multicolumn{2}{c}{\textit{Unheard}}\\
    Model           & {SI-SDR $\uparrow$} & {STFT $\downarrow$} & {SI-SDR $\uparrow$} & {STFT $\downarrow$} \\    \midrule
    Move2Hear                       & \textbf{3.50} & \textbf{0.291} & \textbf{2.33} & \textbf{0.333} \\
    Move2Hear w/o $f^R$      & 2.64  & 0.320 & 1.57  & 0.361 \\
    Move2Hear w/o $V_t$          & 3.32 & 0.300 & 2.12 & 0.343 \\
    Move2Hear w/o $\pi^\mathcal{N}$ & 2.64  & 0.318 & 1.88  & 0.347 \\
    Move2Hear w/o $\pi^\mathcal{Q}$ & 3.08  & 0.304 & 1.99 & 0.343 \\
    \bottomrule
  \end{tabular}
  }
  \vspace*{-0.25cm}
  \caption{Ablation of our Move2Hear model on Far-Target AAViSS}. 
   \label{table:far_target_2source_ablation}
\end{table}


\begin{figure}[t]
    \centering
    \begin{subfigure}[b]{0.7\linewidth}
    \centering
    \includegraphics[width=\linewidth]{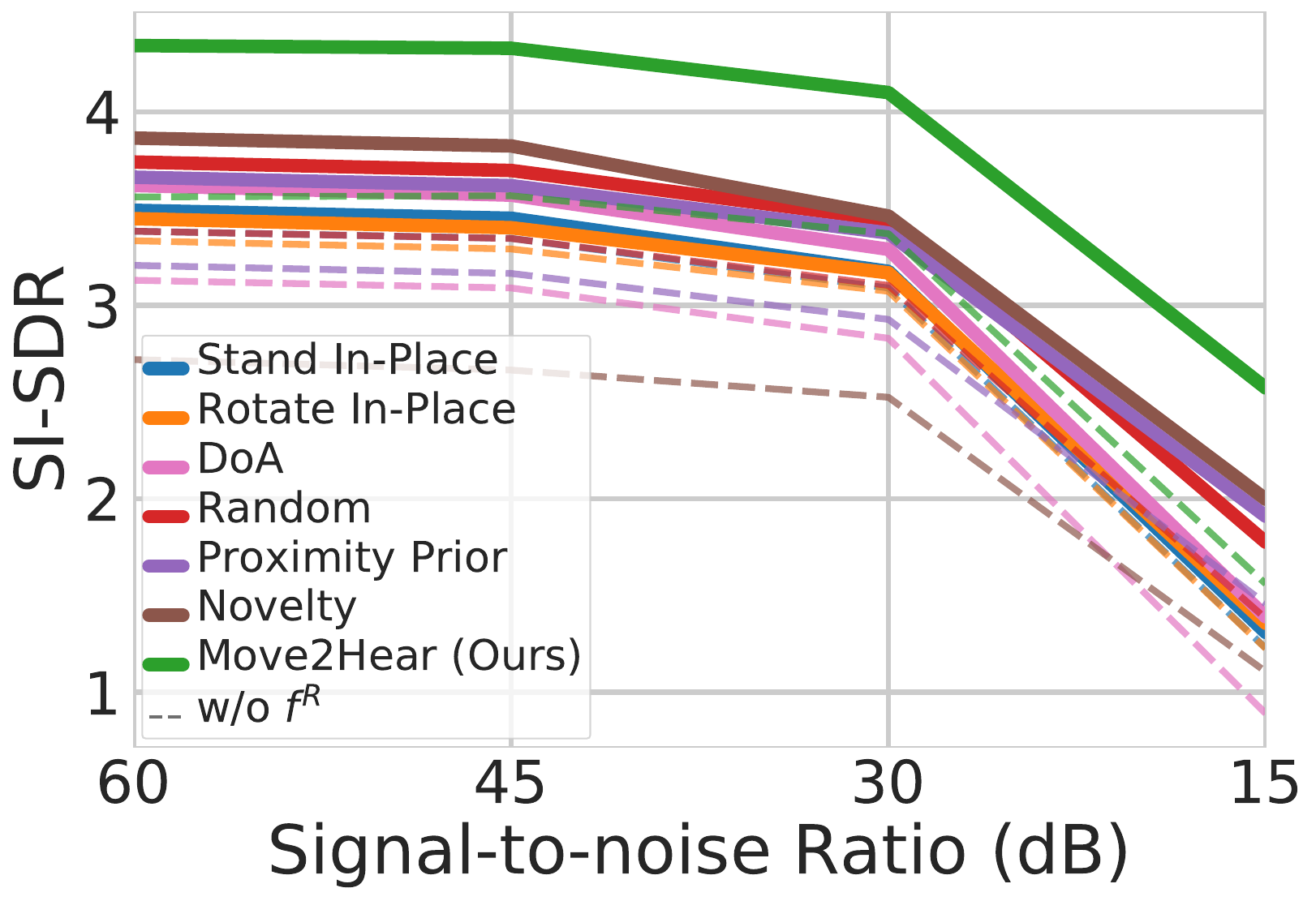}
    \vspace{-0.55cm}
    \caption{Heard sounds}
    \vspace{+0.2cm}
    \label{fig:nearTgt_noise_heard}
    \end{subfigure}\hfill
    \begin{subfigure}[b]{0.7\linewidth}
    \centering
    \includegraphics[width=\linewidth]{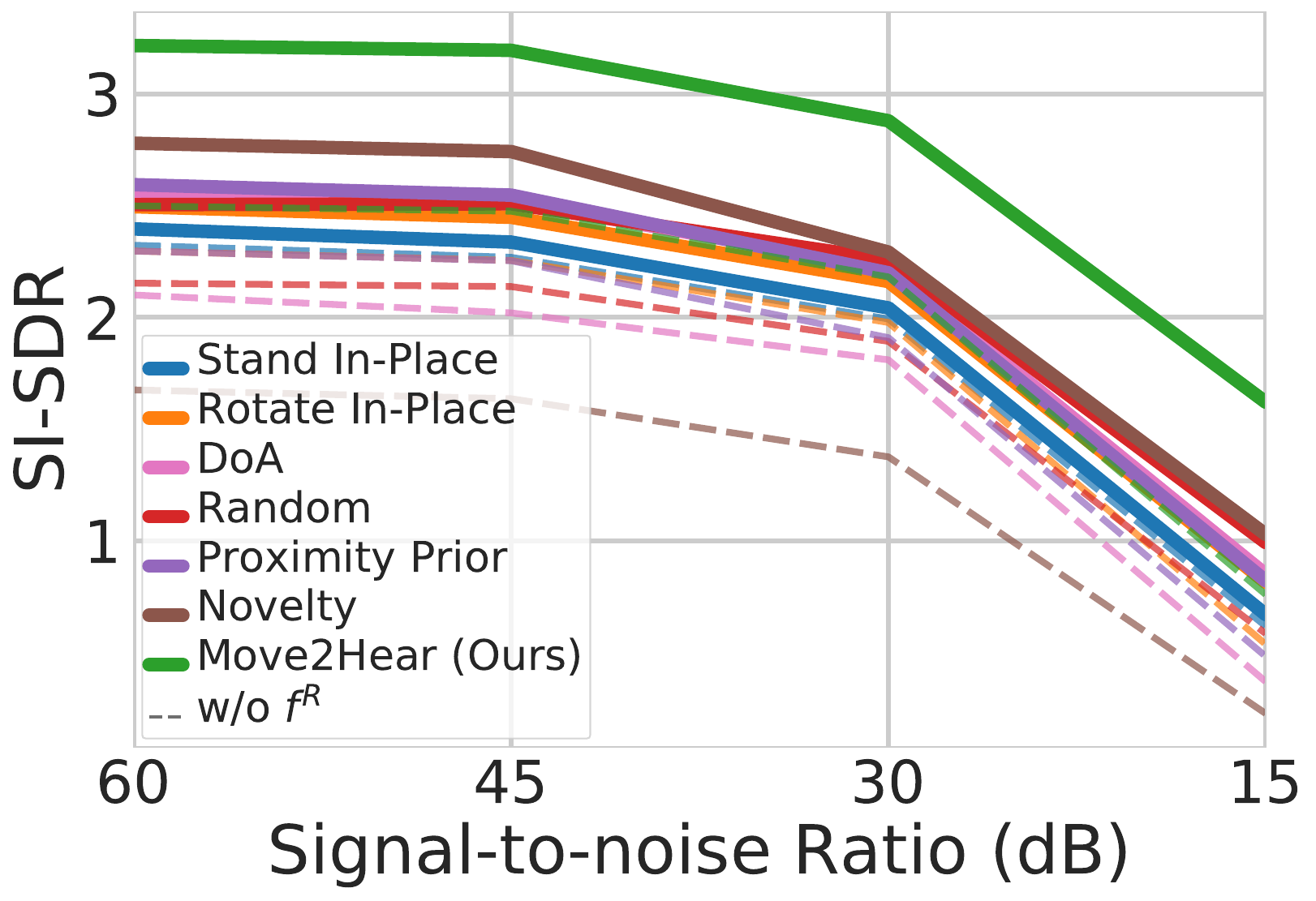}
    \vspace{-0.55cm}
    \caption{Unheard sounds}
    \vspace{+0.2cm}
    \label{fig:nearTgt_noise_unheard}
    \end{subfigure}
    \vspace*{-0.3cm}
\caption{Models' robustness to various levels of noise in audio.}
    \label{fig:noise}
    \vspace{-0.3cm}
\end{figure}


\subsection{Model Analysis}\label{subsec:model_analysis}
\vspace{-0.1cm}
\paragraph{Ablations.} 
In Table~\ref{table:far_target_2source_ablation} we ablate the components 
of our model.
We see that our acoustic memory refiner ($f^R$) plays an important role in the overall performance. $f^R$ promotes stable, improved predictions by informing the policy of the separation quality changes. The vision component $V_t$ is critical as well since $V_t$ helps the agent to avoid obstacles, to reach the target, and to reason about the visible 3D scene.

Please see Supp.~for \textbf{additional analysis} showing that 1) the composite policy for \emph{far-target} is essential for best performance, 
, 2) source types affect task difficulty, 3) Move2Hear retains its benefits even with a SOTA passive separation backbone, 4) 
Move2Hear’s separation quality does not degrade with different intra-source distances,  
and 5) our model helps audio-visual navigation in the presence of distractor sounds.

\vspace{-0.1in}
\paragraph{Noisy Audio.}
We analyze our model's robustness to audio noise using standard noise models~\citep{published_papers/29504406, takeda2017unsupervised} in the 
\emph{near-target} setting (Fig.~\ref{fig:noise}). Our model's gains over the baselines persist even with increasing microphone noise levels for both \emph{heard} and \emph{unheard} sounds. 
The plot also shows the positive impact of our memory refiner $f^R$ (dashed lines); all models decline without it.\footnote{Note that evaluating noisy odometry and actuation is not supported by SoundSpaces since RIRs are available only on the discrete grid.}

\vspace{-0.3cm}
\paragraph{Number of Audio Sources.}
Next we test how our model generalizes to more than one distractor sound. Fig.~\ref{fig:performance_vs_model_target_3_source} shows the results for the \emph{near-target} task using $k=3$ audio sources per episode. Our model generalizes better than the rest of the baselines and maintains its advantage.

\begin{figure}[t]
\centering 
\includegraphics[width=1.\linewidth]{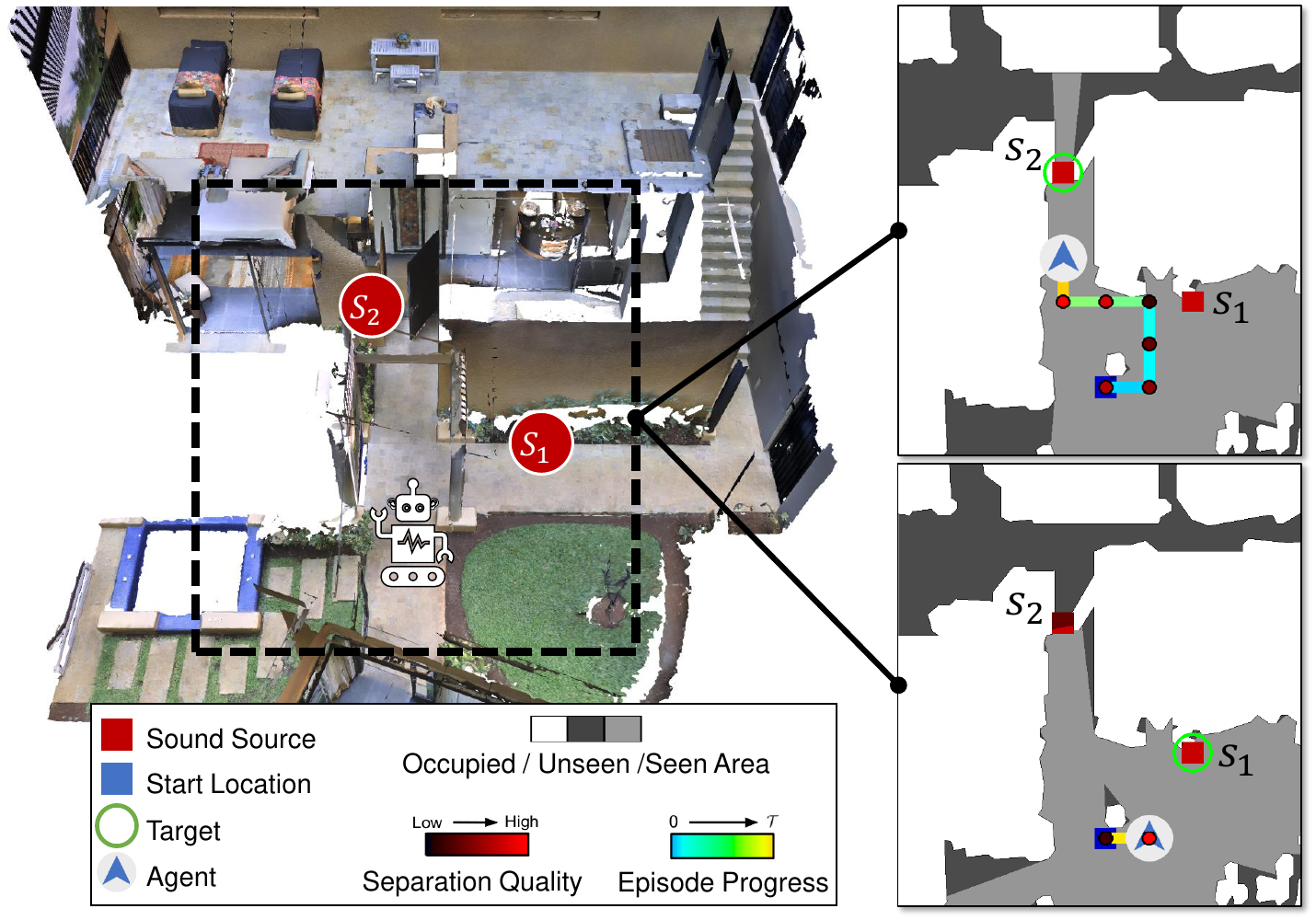}
\vspace*{-0.2in}
\caption{Example movements by our Move2Hear model. Our model takes advantage of the visible 3D structure to actively improve its separation quality of a target audio (see text for details).}
\label{fig:qual}
\end{figure}

\vspace{-0.3cm}
\paragraph{Qualitative Results.}
In Fig.~\ref{fig:qual}, our Move2Hear agent is placed in a scene with two audio sources as possible targets. Our model exhibits an intriguing behavior that takes advantage of the visible 3D structures.  When  $S_1$ is the target, it takes the minimum steps to go around the column to put itself in the acoustic shadow of the column in relation to $S_2$, thus dampening its signal. However, when $S_2$ is the target, it decides to move into the corridor closer to $S_2$, putting the wall between itself and $S_1$.

\vspace{-0.3cm}
\paragraph{Failure Cases.}
Common failure cases for \emph{near-target} involve the agent having limited freedom of movement due to complex surrounding geometry and when any translational motion takes it towards the distractor(s) thus incurring a 
high loss in quality. For \emph{far-target}, the agent sometimes lacks a direct path to the target due to cluttered surroundings.

%% file: sections/conclusion.tex
\vspace*{-0.05in}
\section{Conclusion}
\vspace*{-0.05in}

We introduced the AAViSS task, where agents must move around using both sight and sound to best listen to a desired target object.  Our Move2Hear model offers promising results, consistently outperforming alternative exploration/navigation motion policies from the literature, as well as strong baselines. In future work, we aim to extend our model to account for non-periodic sounds, e.g., with new forms of sequential memory, and to investigate sim2real transfer of the learned policies.

\noindent 
\textbf{Acknowledgements:} UT Austin is supported in part by DARPA L2M and the IFML NSF AI  Institute.  K.G. is paid as a Research Scientist by Facebook AI.

%% file: sections/supp.tex
\section{Supplementary Material}
In this supplementary material we provide additional details about:
\begin{itemize}
    \item Video (with audio) for qualitative assessment of our task setup and agent's performance (Sec.~\ref{sec:video}), as referenced in `3D Environment and Audio-Visual Simulator' of Sec.~\ref{sec:task} in the main paper.
    \item Noisy audio experiment for the \emph{far-target} task (Sec.~\ref{sec:supp_noise}).
    \item Experiment to show the effect of the minimum inter-source distance on separation quality (Sec.~\ref{sec:supp_intersource_dist}), as noted in Sec.~\ref{subsec:model_analysis}
    of the main paper.
    \item Experiment to demonstrate the importance of the \emph{composite policy} for \emph{far-target} separation (Sec.~\ref{sec:supp_composite_policy}), as mentioned in Sec.~\ref{subsec:model_analysis}
    of the main paper.
    \item Experiment to show how the separation quality varies across all possible type of configurations of combining the target and the distractor sources.
    (Sec.~\ref{sec:supp_diff_scenarios}), as noted in Sec.~\ref{subsec:model_analysis} of the main paper.
    \item Experiment to show how our Move2Hear approach maintains its benefits even when using 
    a SOTA passive audio separation backbone  (Sec.~\ref{sec:supp_mmdensenet}), as noted in Sec.~\ref{subsec:model_analysis} of the main paper.
    \item Experiment to show the effect of using waveform-level audio quality metrics like SNR as the RL reward on the separation performance (Sec.~\ref{sec:supp_snr_reward}).
    \item Experiment to show how audio-visual navigation with distractor sources benefits from active audio-visual source separation (Sec.~\ref{sec:supp_avnav_distractors}) as mentioned in Sec.~\ref{subsec:model_analysis}
    in the main paper. 
    \item Evaluation metric definitions for evaluating source separation quality (Sec.~\ref{sec:supp_metric_definitions}).
    \item Additional baseline details for reproducibility (Sec.~\ref{sec:supp_baselines}).
    \item Implementation details (Sec.~\ref{sec:supp_implementation_details}), as promised in `Experimental Setup' of Sec.~\ref{sec:experiments}
    in the main paper.
\end{itemize}

\begin{figure*}[t]
    \centering
    \begin{subfigure}[b]{0.45\linewidth}
    \centering
    \includegraphics[width=\linewidth]{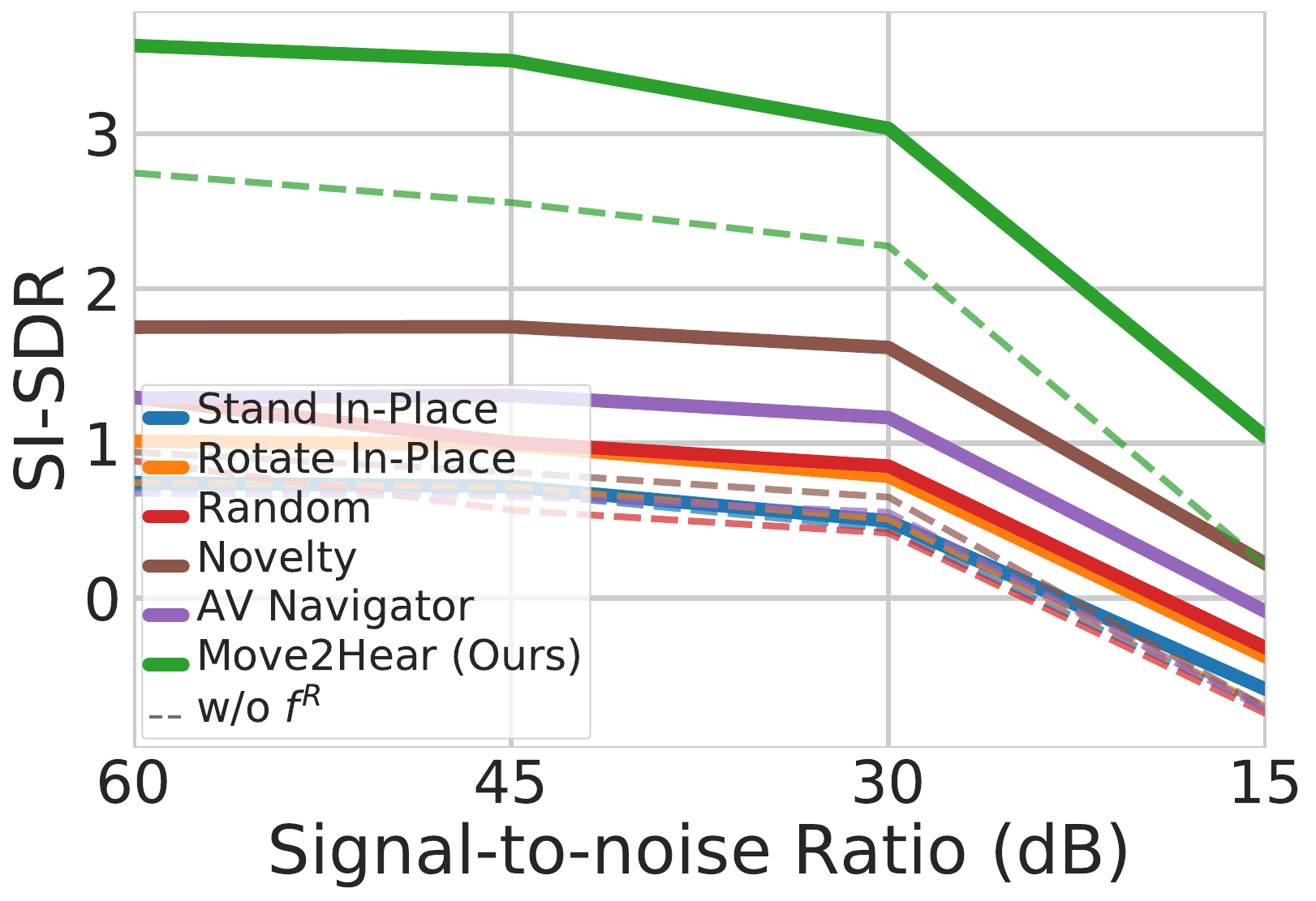}
    \vspace{-0.55cm}
    \caption{Heard sounds}
    \vspace{+0.2cm}
    \label{fig:farTgt_noise_heard}
    \end{subfigure}
    \begin{subfigure}[b]{0.45\linewidth}
    \centering
    \includegraphics[width=\linewidth]{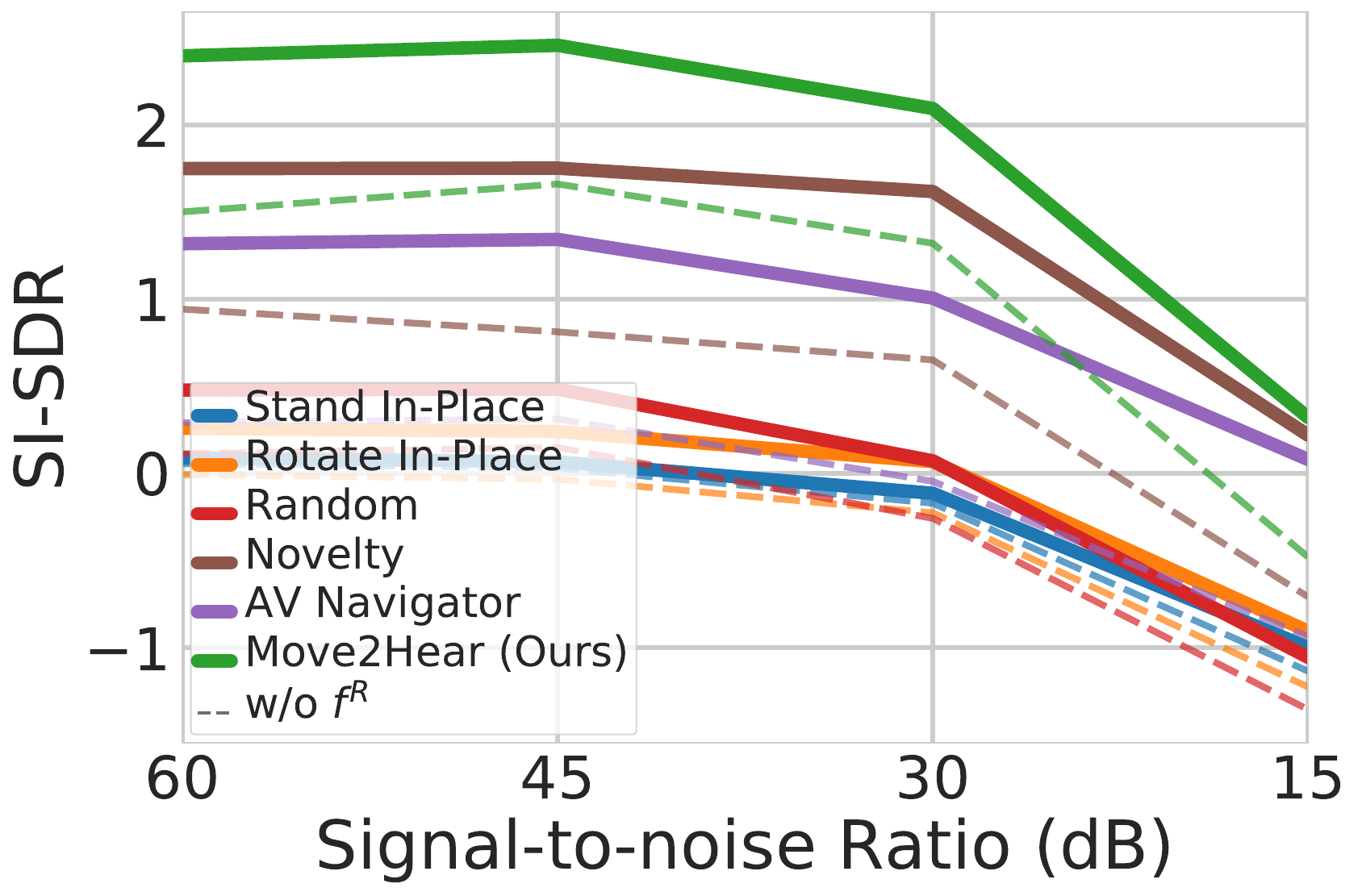}
    \vspace{-0.55cm}
    \caption{Unheard sounds}
    \vspace{+0.2cm}
    \label{fig:farTgt_noise_unheard}
    \end{subfigure}
\caption{Models' robustness for various levels of noise in audio for \emph{far-target}. Higher SI-SDR is better.}
    \label{fig:performance_vs_noise_far_target_2_source}
\end{figure*}

\subsection{Supplementary Video}\label{sec:video}
The supplementary video, available at \url{http://vision.cs.utexas.edu/projects/move2hear}, demonstrates the Active Audio-Visual Source Separation task with the SoundSpaces~\citep{chen2020soundspaces} audio simulation setup and shows the comparison between our proposed model and the baselines as well as qualitative results for both \emph{near-target} and \emph{far-target}.  Please listen with headphones to hear the binaural audio correctly. 

\subsection{Noisy Audio for Far-Target}\label{sec:supp_noise}
So far, we tested our model's robustness against standard microphone noise~\citep{published_papers/29504406, takeda2017unsupervised} in the \emph{near-target} task setup (see Fig.~4 above). Here, we show the parallel experiment for the \emph{far-target} task setup. Fig.~\ref{fig:performance_vs_noise_far_target_2_source} shows the results. Our model is able to maintain its performance gain in the \emph{far-target} setup over all other models even for very high levels of noise in both the \emph{heard} and \emph{unheard} settings. In addition, we see that as in the case of \emph{near-target}, our acoustic memory refiner module $f^R$ again plays an important role in providing additional robustness against noisy audio; all models perform worse without it.

\subsection{Minimum Inter-Source Distance}\label{sec:supp_intersource_dist}
\begin{figure*}[t]

\centering 
\begin{subfigure}[b]{0.45\linewidth}
\centering
\includegraphics[width=\linewidth]{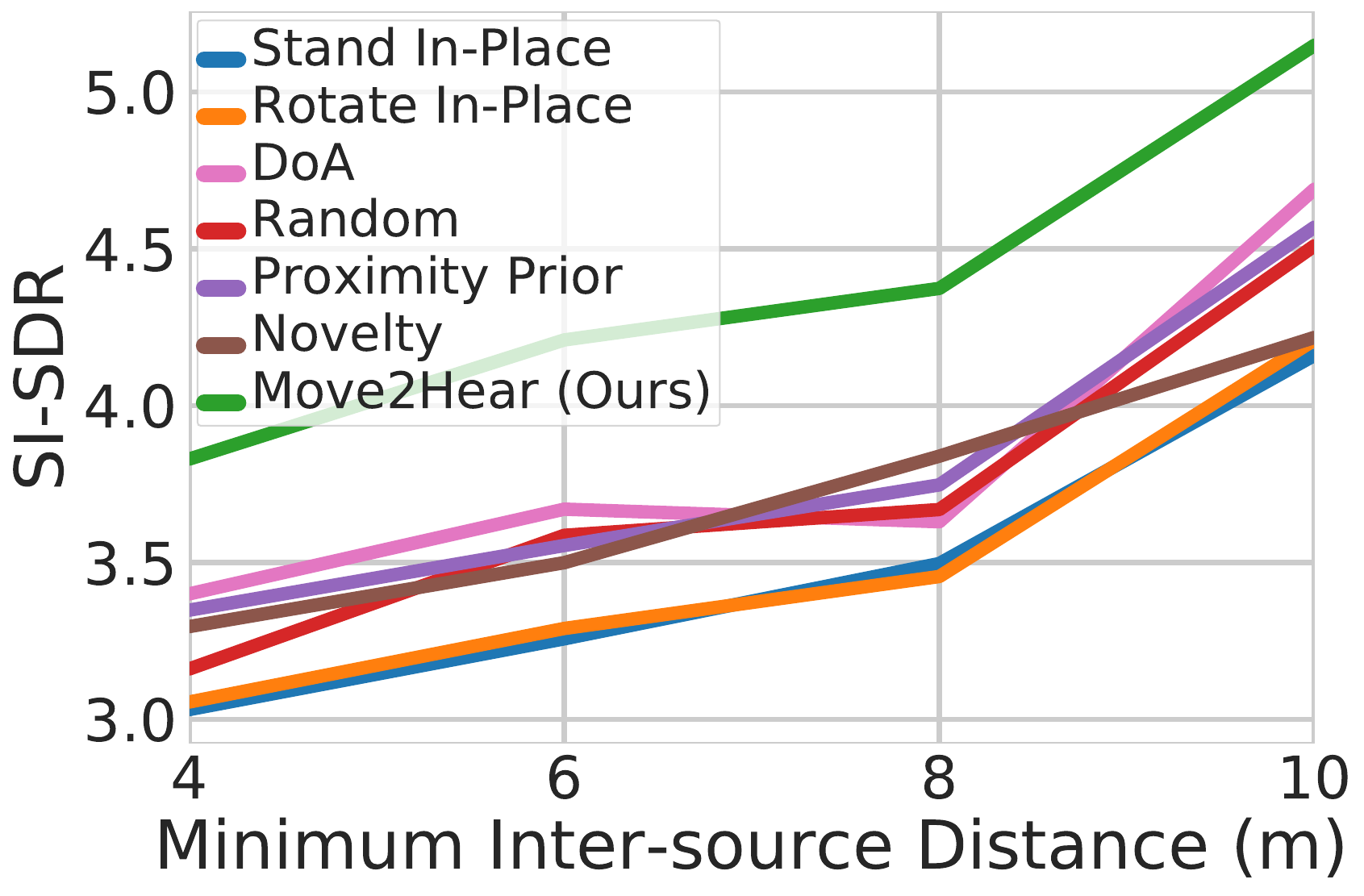}
\vspace{-0.55cm}
\caption{Near-target heard}
\vspace{0.2cm}
\end{subfigure}
\begin{subfigure}[b]{0.45\linewidth}
\centering
\includegraphics[width=\linewidth]{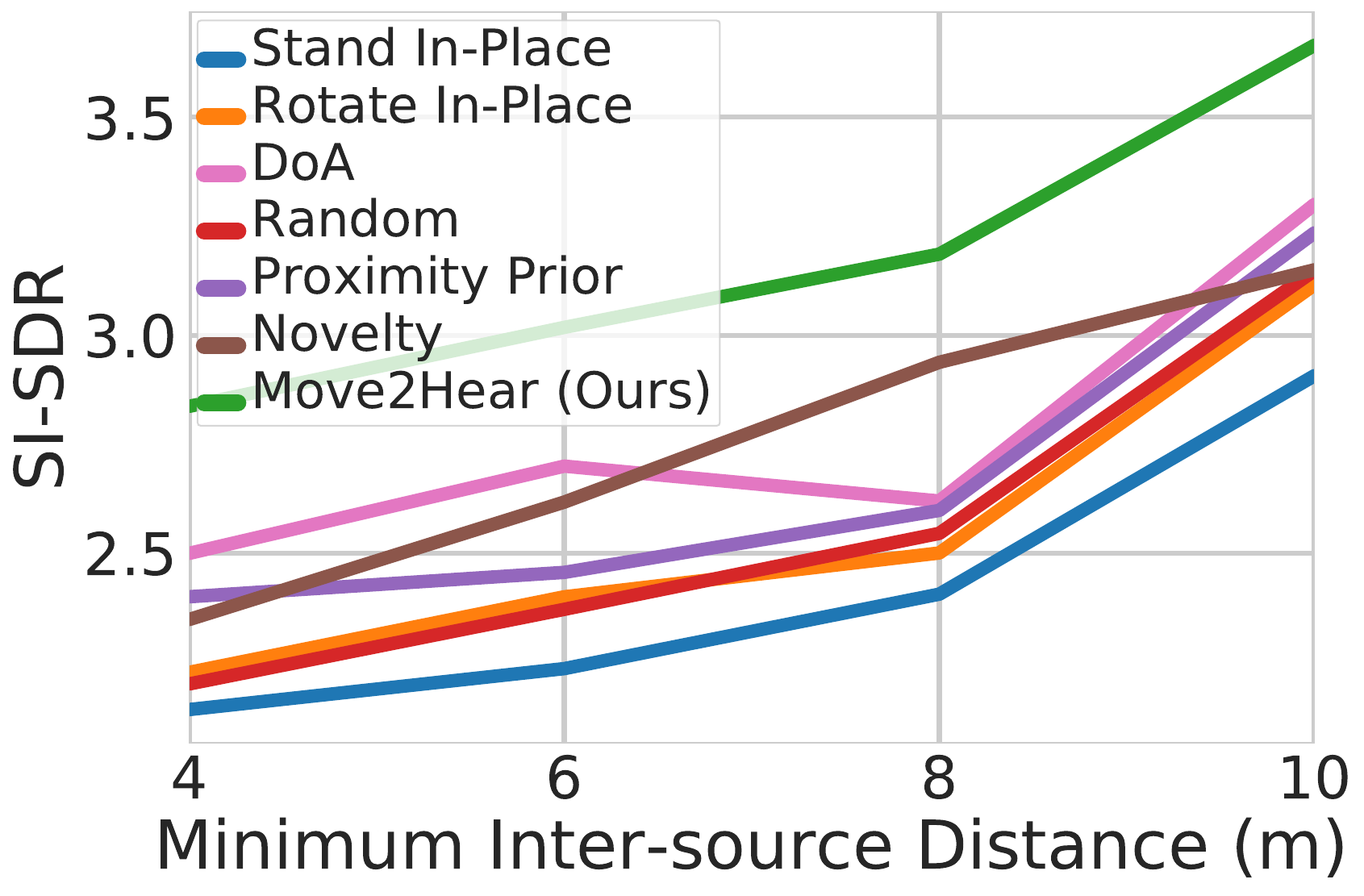}
\vspace{-0.55cm}
\caption{Near-target unheard}
\vspace{0.2cm}
\end{subfigure}

\centering 
\begin{subfigure}[b]{0.45\linewidth}
\centering
\includegraphics[width=\linewidth]{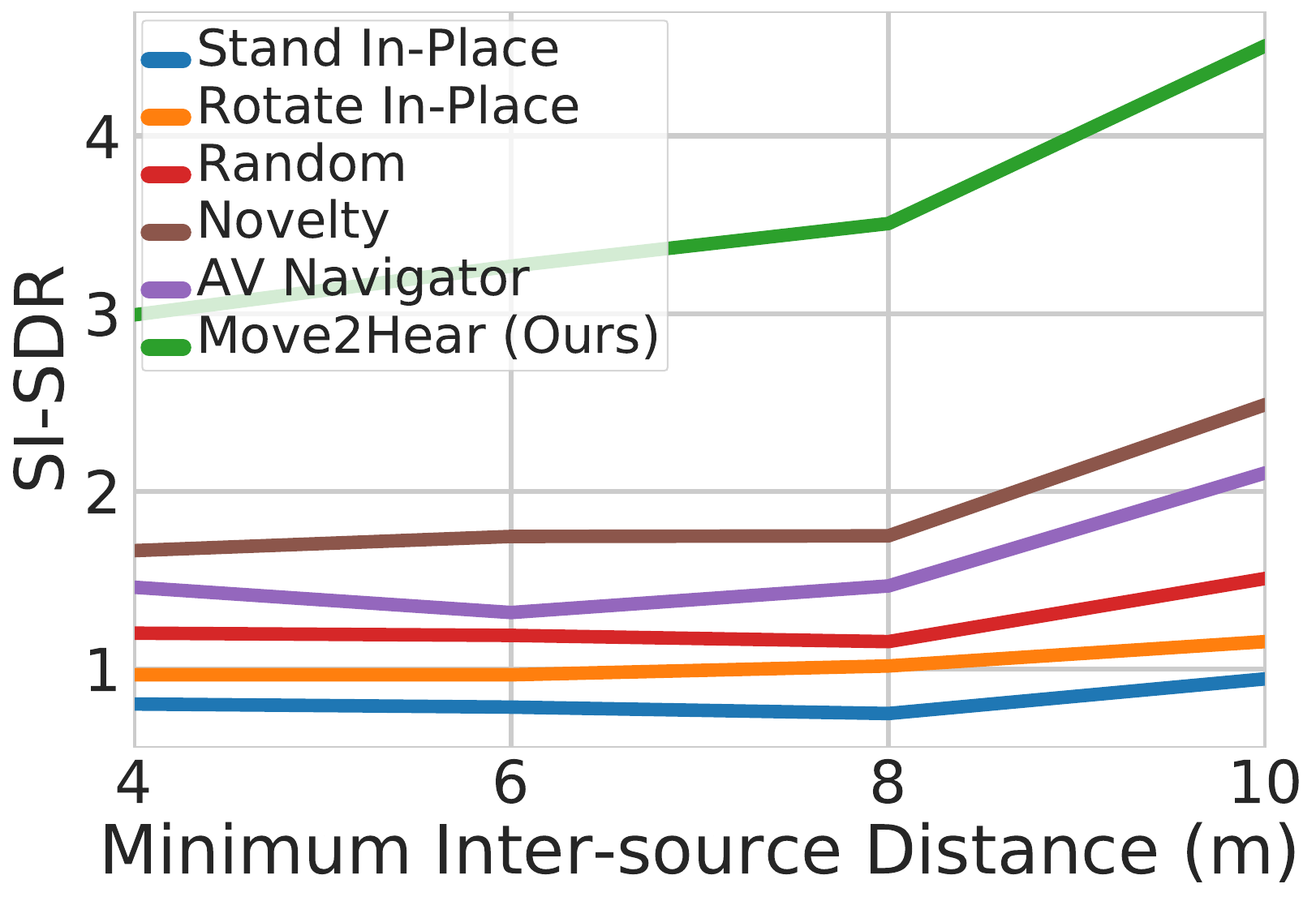}
\vspace{-0.55cm}
\caption{Far-target heard}
\vspace{0.2cm}
\end{subfigure}
\begin{subfigure}[b]{0.45\linewidth}
\centering
\includegraphics[width=\linewidth]{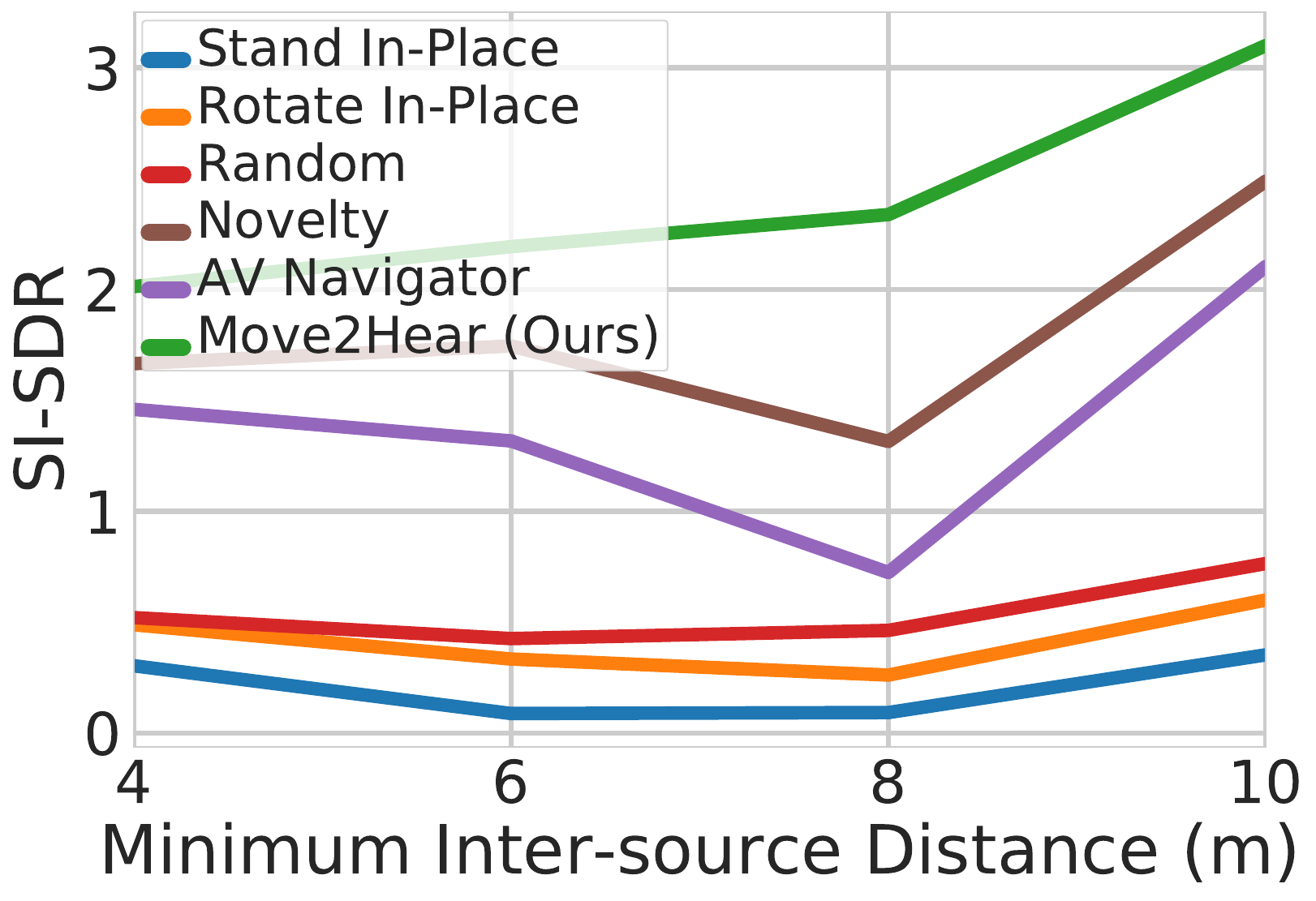}
\vspace{-0.55cm}
\caption{Far-target unheard}
\vspace{0.2cm}
\end{subfigure}

\caption{Models' robustness to  various inter-source distances.  Higher SI-SDR is better.}
\label{fig:performance_vs_S2SMinDist_far_target_2_source}
\end{figure*}

We investigate the effect of the minimum inter-source distance (`Experimental Setup' in Sec.~\ref{sec:experiments} in main) on the separation performance of our model. This minimum distance is applied to all audio source pairs in every episode. Fig.~\ref{fig:performance_vs_S2SMinDist_far_target_2_source} shows the results with both \emph{heard} and \emph{unheard} sounds. 

For both settings, our Move2Hear model outperforms all baselines by a significant margin. This shows that even in the challenging setting where the target and the distractor are quite close to each other (i.e., the maneuverability space around the target is reduced and the clash between the sounds of the target and the nearby distractor can be high), our model can still actively move around to effectively improve its separation quality.

\subsection{Importance of Composite Policy}\label{sec:supp_composite_policy}
Our composite policy switches control between the navigation policy $\pi^\mathcal{N}$ and the quality policy $\pi^\mathcal{Q}$ based on the policy switch time  $\mathcal{T}^\mathcal{N}$, where $\mathcal{T}^\mathcal{N}$ is selected based on performance in the validation split.
Inspired by recent work in composite policy blending for complex multi-task robot learning~\citep{6630814, devin2017learning, haarnoja2018composable, su2018learning},
this approach helps the agent deal with the challenges posed by the far-target task.
When the agent is too far from the target audio location, the audio signal could be too weak and unreliable for $\pi^{\mathcal{Q}}$ to perform reasonably. Hence, $\pi^{\mathcal{N}}$ brings the agent to an area with a stronger signal and then passes control to $\pi^{\mathcal{Q}}$, which is expert in moving to improve $\ddot{M}^G$.

\begin{figure}[t]
\centering 
\includegraphics[width=0.4\textwidth]{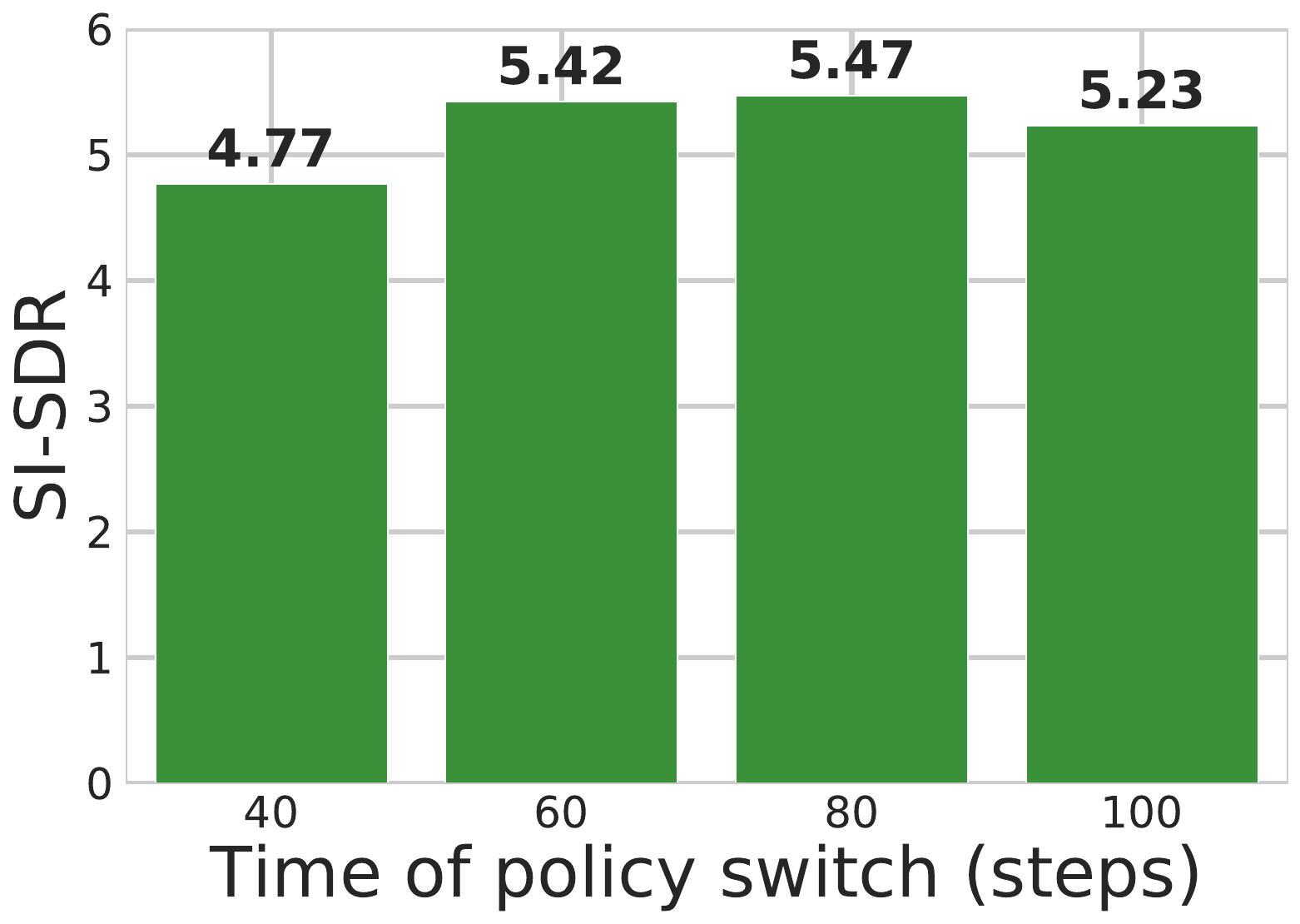}
\caption{Effect of policy switch time in composite policy on separation performance for \emph{far-target}.}
\label{fig:performance_vs_TimeOfPolicySwitch_far_target_2_source}
\vspace*{-0.3in}
\end{figure}

Fig.~\ref{fig:performance_vs_TimeOfPolicySwitch_far_target_2_source} shows Move2Hear's separation performance on the validation data for different values of the policy switch time $\mathcal{T}^\mathcal{N}$ in the \emph{far-target} and \emph{heard} setting. 
Switching over from the navigation policy $\pi^\mathcal{N}$ to the quality improvement policy $\pi^\mathcal{Q}$ very early negatively affects our model's performance as it does not allow the agent to be close enough to the source for $\pi^\mathcal{Q}$ to make successful fine-grained movements for further improvement in separation quality. On the other hand, if there is no switching at all ($\mathcal{T}^\mathcal{N}=100$), the agent suffers from not leveraging $\pi^\emph{Q}$'s ability to take it to ``sweet spots'' in the vicinity of the target where the target audio can be separated even better. Overall, the composite policy is beneficial for best results, and we see the model is not overly sensitive to the switch point.

\subsection{Separation in Different Scenarios}\label{sec:supp_diff_scenarios}
Separating speech (S) from a mixture of speeches is the most difficult, and extracting music (M) from background (B) is the easiest for Move2Hear and two of our strongest baselines in the \emph{near-target} scenario, Novelty~\citep{bellemare2016unifying} and Proximity Prior. For the SI-SDR scores, refer to Table~\ref{table:near_target_2source_diff_audio_pairs}.

\begin{table}[!h]
\vspace{-0.2cm}
  \centering
    \scalebox{0.85}{
    \setlength{\tabcolsep}{6pt}
    \begin{tabular}{|c|l|c|c|c|c}
    \toprule
    & Model           & S vs. S & S vs. M & S vs. B & M vs. B\\    \midrule
    \multirow{3}*{\rotatebox{90}{\footnotesize Heard}} 
    & Proximity Prior & 3.42  & 5.07 & 3.38  & 6.82 \\
    & Novelty~\citep{bellemare2016unifying} & 3.56  & 4.84 & 3.85  & 5.83 \\
    & Move2Hear (Ours)             & \textbf{4.04} & \textbf{5.50} & \textbf{4.58} & \textbf{6.95} \\
    \midrule
    \multirow{3}*{\rotatebox{90}{\footnotesize Unheard}} 
    & Proximity Prior         & 2.53  & 2.78 & 2.69  & 3.36 \\
    & Novelty~\citep{bellemare2016unifying}        & 2.82  & 3.19 & 3.25  & 3.82 \\
    & Move2Hear (Ours)    & \textbf{3.08} & \textbf{3.31} & \textbf{3.60} & \textbf{4.15} \\
    \bottomrule
  \end{tabular}
  }
  \vspace*{-0.15cm}
  \caption{SI-SDR performance in different separation scenarios on \emph{near-target}.
  } \label{table:near_target_2source_diff_audio_pairs}
\end{table}

\subsection{Comparison with SOTA passive separation model.}\label{sec:supp_mmdensenet}
Passive audio(-visual) separation is distinct from AAViSS in that it 1) assumes access to pre-recorded audio/video, 2) has no provision for sensor motion to improve separation, and 3) doesn't extract the target latent (monaural) audio (`Passive Audio(-Visual) Source Separation' of Sec. 2 in main). Furthermore, advances in passive audio separation models are orthogonal to our contribution. We demonstrate this by replacing the audio network backbone with the passive SOTA MMDenseNet~\citep{takahashi2017multi} On SI-SDR and heard (unheard) setting, our active model still outperforms Stand In-Place in both \emph{near-} and \emph{far-target} settings (Table~\ref{table:nrAndFr_target_2source_mmDnsNt}).

\begin{table}[!h]
  \centering
    \scalebox{0.85}{
    \setlength{\tabcolsep}{2pt}
    \begin{tabular}{lcc|cc}
    \toprule
                    &   \multicolumn{2}{c|}{Near-Target} &  \multicolumn{2}{c}{Far-target}\\
    Model           & \emph{Heard} & \emph{Unheard} & \emph{Heard} & \emph{Unheard} \\    \midrule
    Stand In-Place      & 7.10  & 4.43 & 3.45  & 1.90 \\
    Move2Hear (Ours) & \textbf{7.98} & \textbf{5.38} & \textbf{6.84} & \textbf{4.57}\\ 
    \bottomrule
  \end{tabular}
  }
  \vspace*{-0.15cm}
  \caption{Effect of using a stronger model like MMDenseNet~\citep{takahashi2017multi} for passive separation on SI-SDR performance.}\label{table:nrAndFr_target_2source_mmDnsNt}
\end{table}

\subsection{SNR as RL reward}\label{sec:supp_snr_reward}
In our approach, we used the source separation error in the formulation of the agent's reward. Here, we explore an alternative option for the reward formulation by replacing the separation error with SNR (signal to noise ratio) to capture the improvement in quality of the waveform-level of the separated audio.  We find that this alternative RL reward doesn't improve the separation performance. On the contrary, when using SNR as a reward with Move2Hear we see a relative degradation in SI-SDR by 5.1\% and 2.2\% in near-target and heard/unheard setting in comparison to the original reward formulation. Further, SNR increases training time by 2.2x due to the needed inverse-STFT calculations. Our separation reward leads to better performance and faster training.

\begin{table}[t]
  \scalebox{0.85}{
  \centering
  \setlength{\tabcolsep}{4pt}
    \begin{tabular}{l SS|SS}
    \toprule
    &   \multicolumn{2}{ c| }{\textit{Heard}} &  \multicolumn{2}{ c }{\textit{Unheard}}\\
    Model  & {SPL ($\uparrow$)} & {SR ($\uparrow$)} & {SPL ($\uparrow$)} & {SR ($\uparrow$)}\\
    \midrule
    Random & 3.1 & 6.4 & 3.1 & 6.4  \\
    Move Forward & 1.1 & 1.1 & 1.1 & 1.1 \\
    \midrule
    \multicolumn{5}{l}{\textbf{Speaker-Target}}\\ 
    Gan et al.~\citep{gan2019look} & 5.2 & 12.3 & 4.3 & 10.3  \\
    AV Navigator~\citep{chen2020soundspaces} & 33.5 & 49.1 & 32.4 & 47.0 \\
    Move2Hear [$\pi^\mathcal{N}$ + $Stop$] (\textbf{Ours}) & \textbf{56.0} & \textbf{70.0} & \textbf{51.4} & \textbf{66.0}\\
    \midrule
    \multicolumn{5}{l}{\textbf{Standard Split}}\\ 
    Gan et al.~\citep{gan2019look}  & 4.3 & 10.0 & 4.9 & 11.1 \\
    AV Navigator~\citep{chen2020soundspaces}  & 0.9 & 1.5 & 1.1 & 1.6 \\
    Move2Hear [$\pi^\mathcal{N}$ + $Stop$] (\textbf{Ours}) & \textbf{54.9} & \textbf{70.3} & \textbf{52.2} & \textbf{68.5}\\
  \end{tabular}
  }
  \caption{Audio-visual navigation with distractors. Higher SPL and SR are better.
  }\label{table:pure_nav_2source_supp}
\vspace{-0.3cm}
\end{table}

\subsection{Audio-visual Navigation with Distractors}\label{sec:supp_avnav_distractors}
While our main goal is source separation, we find that as a byproduct, our model can benefit AV navigation in the presence of cluttered sounds.  Whereas existing models trained to navigate to a source are naturally confused by distractors, our $\pi^\mathcal{N}$ navigation policy (augmented with a \emph{Stop} action) can successfully ignore them to more rapidly find a target source. 
To illustrate this, we use the \emph{far-target} dataset and we compare our $\pi^\mathcal{N}$ policy to the following models in terms of navigation performance:
\begin{itemize}
    \item \textbf{Random:} an agent that is similar to the random baseline in AAVISS but samples random actions from the augmented action space instead.
    \item \textbf{Move Forward:} an agent that always moves forward unless faced with an obstacle, then it turns right. This is a common baseline employed in the visual navigation literature~\citep{habitat19iccv, chen2020soundspaces}.
    \item \textbf{AV Navigator~\citep{chen2020soundspaces}}: this is the same baseline model we used for the \emph{far-target} setting but evaluated here for navigation performance.
    \item \textbf{Gan et al.~\citep{gan2019look}}: this approach trains two supervised models using the binaural audio input, one for predicting the target location and the other for predicting a \emph{Stop} action. During navigation, the method of Gan et al.~\citep{gan2019look} uses egocentric depth images to build an occupancy map of the environment and plans a path to the predicted location using a metric planner. We set the target location prediction frequency to every 20 steps of navigation on the basis of validation.
\end{itemize}

All models are evaluated using standard navigation metrics: success rate (SR) and success rate weighted by path length (SPL).

Table~\ref{table:pure_nav_2source_supp} shows the results. On the \emph{Standard Split} when the target and distractor types intersect, both \cite{chen2020soundspaces} and \citep{gan2019look} are overwhelmed by the mixed audio and show poor navigation performance. We observe that per-step prediction for the Gan et al.~model yields a very reactive navigation policy in our setup, which leads to low navigation performance. On an easier split where the target is always of speaker type and the distractors are never other speakers (\emph{Speaker-Target}), the learned baselines fare better. However, our model outperforms all baselines by a substantial margin in both setups, showing the positive impact of using separated audio for navigation with distractor sounds.

\subsection{Metric Definitions}~\label{sec:supp_metric_definitions}
Next we elaborate on the metric definitions (`Evaluation' of Sec.~\ref{sec:experiments} in main).

\begin{enumerate}
    \item \textbf{STFT distance --} The Euclidean distance between the ground-truth and predicted complex monaural spectrograms,
    \begin{equation*}
        \mathcal{D}_{\{STFT\}} = ||\boldsymbol{\ddot{M}}^{G} - \boldsymbol{M}^{G}||_{2}.
    \end{equation*}
    \item \textbf{SI-SDR~\citep{8683855} --} We use a fast implementation from the nussl~\citep{nussl} library to measure the source-to-distortion ratio (SDR) of the predicted monaural waveforms in dB in a scale-invariant (SI) manner.
\end{enumerate}

\subsection{Baselines}\label{sec:supp_baselines}
We provide the following additional details about the baselines (`Baselines' of Sec.~\ref{sec:experiments} in main) for reproducibility.
\begin{itemize}
    \item \textbf{DoA:} To face the audio target, this agent starts rotating to the right from its initial pose until it finds an orientation that allows it to move to a neighboring node. Once it has moved to a neighboring node, it rotates twice to face the agent and make its first prediction.
    \item \textbf{Proximity Prior:} Whenever this agent tries to cross the boundary of the circle within which it is supposed to stay, it is forced to randomly choose an action from $\{TurnLeft, TurnRight\}$ by the simulation platform.
    \item \textbf{Novelty~\citep{bellemare2016unifying}:} this agent is rewarded on the basis of the  novelty of states visited. Each valid node of the SoundSpaces~\citep{chen2020soundspaces} grids is considered to be a unique state. When an agent visits any such node, the count for
    that state is incremented. The novelty reward is given by:
    \begin{equation}
        r_t = \frac{1}{\sqrt{n_s}},
    \end{equation}
    where $n_s$ is the visitation count of state $s_t$.
    \item \textbf{AV Navigator~\citep{chen2020soundspaces}:} this navigation agent uses a visual and an audio encoder for input feature representation and an actor-critic policy network for predicting actions to navigate to the target source. While the visual encoder takes RGB images as input, the audio encoder takes the mixed binaural spectrogram concatenated with the target class label as an extra channel as input. Thus, the audio input space is the same as the input space of $f^B$ (Eqn.~\ref{eq:f_B} in Sec.~\ref{approach:audio_net} of main). Following typical navigation rewards~\citep{chen2020soundspaces, habitat19iccv}, we reward the agent with +10 if it succeeds in reaching the target source and executing the Stop action there, plus an additional reward of +0.25 for reducing the geodesic distance to the target and an equivalent penalty for increasing it. Finally, we issue a time penalty of -0.01 per executed action to encourage efficiency.
\end{itemize}

\subsection{Implementation Details}\label{sec:supp_implementation_details}

Next we provide further implementation details including the network architecture details.

\subsubsection{Monaural Audio Preprocessing}
For all our experiments, we sample 1-second-long monaural clips (`Experimental Setup' of Sec.~\ref{sec:experiments} in main) at 16kHz and ensure that the sampled clips have a higher average power than the full audio clip that they are sampled from. This helps us prevent the sampling of a large amount of mostly silent raw audio data. The sampled waveforms are further encoded using the standard 32-bit floating point format and normalized to have the same average power of 1.2 across the whole dataset.

\subsubsection{Audio Spectrogram}
To generate an audio spectrogram, we compute the Short-Time Fourier Transform (STFT) with a Hann window of length 63.9ms, hop length of 32ms, and FFT size of 1023. 
This results in complex spectrograms of size $512 \times 32 \times C$, where $C$ is the number of channels in the source audio ($C$ is 1 for monaural and 2 for binaural audio). 
For all experiments, we take the magnitude of the spectrogram, compute its natural logarithm after adding 1 to all its elements for better contrast~\citep{gao2019visual-sound, gao2019co}, and reshape it to $32 \times 32 \times 16C$ by taking slices along the frequency dimension and concatenating them channel-wise to improve training speed. For all cases where the target audio class needs to be concatenated to the spectrogram channel-wise, the concatenation is carried out after slicing.

\subsubsection{CNN Architecture Details}
\paragraph{Binaural Audio Separator.}
The binaural audio separator $f^B$ uses a U-Net style architecture~\citep{RFB15a} (`Binaural Auio Separator' of Sec.~\ref{approach:audio_net} in main). The encoder of the network has 5 convolution layers. Each convolution layer uses a kernel size of 4, a stride of 2 and a padding of 1. 
It is followed by a Batch Normalization~\citep{ioffe2015batch} of $1e^{-5}$ and a leaky ReLU~\citep{DBLP:conf/icml/NairH10, sun2015deeply} activation with a negative slope of 0.2. The number of output channels of the convolution layers are [64, 128, 256, 512, 512], respectively. 
The decoder consists of 5 transpose convolution layers and 1 convolution layer in the end to resize the output from the transpose convolutions to the desired spectrogram dimensions. Each transpose convolution has a kernel size of 4, a stride of 2 and a padding of 1, and is followed by a Batch Normalization~\citep{ioffe2015batch} of $1e^{-5}$ and a ReLU activation~\citep{DBLP:conf/icml/NairH10, sun2015deeply}. The final convolution layer uses a kernel size of 1 and a stride of 1. 

\paragraph{Monaural Audio Predictor.} The monaural audio predictor $f^{M}$ uses the same architecture as $f^B$.

\paragraph{Acoustic Memory Refiner.}
The acoustic memory refiner $f^R$ is a CNN network with 2 convolution layers. Both convolutions use a kernel size of 3, a stride of 1 and a padding of 1. Additionally, the first convolution is followed by a Batch Normalization~\citep{ioffe2015batch} of $1e^{-5}$ and a ReLU activation~\citep{DBLP:conf/icml/NairH10, sun2015deeply}.

\paragraph{Visual Encoder.}
The visual encoder $E^V$ of Move2Hear is a CNN with 3 convolution layers, where the convolution kernel sizes are [8, 4, 3], the strides are [4, 2, 1] and the number of output channels are [32, 64, 32], respectively. Each convolution layer has a ReLU activation~\citep{DBLP:conf/icml/NairH10, sun2015deeply} function. The convolution layers of the encoder are followed by 1 fully connected layer with 512 output units. 
Note that the visual encoders of the AV Navigator~\citep{chen2020soundspaces} and Novelty~\citep{bellemare2016unifying} baselines share the same architecture.

\paragraph{Separated Binaural Encoder.}
Our separated binaural encoder $E^B$ uses the same architecture as $E^V$, except for using a kernel size of 2 in place of 3 for the third convolution.

\paragraph{Policy Network.}
The policy network for Move2Hear, as well as for the AV Navigator~\citep{chen2020soundspaces} and Novelty~\citep{bellemare2016unifying} models, uses a one-layer bidirectional GRU~\citep{NIPS2015_b618c321} with 512 hidden units. The actor and the critic networks consist of one fully connected layer. 

\paragraph{Predicted Monaural Encoder.}
Our predicted monoaural encoder $E^M$ uses the same architecture as $E^B$.
\\\\We use the Kaiming-normal~\citep{He_2015_ICCV} weight initialization strategy for all weight initializations in the network components ($f^B$, $f^R$, $f^M$) of the Target Audio Separator, all feature encoders ($E^V$, $E^B$, $E^M$) of the Active Audio-Visual Controller, and the visual encoder of the AV Navigator~\citep{chen2020soundspaces} and Novelty~\citep{bellemare2016unifying} models.

\subsubsection{Training Hyperparameters}
We pretrain $f^B$ and $f^M$ by randomly sampling a maximum of 30K data samples per training scene (Sec.~\ref{approach:training} in main). We optimize the loss functions $\mathcal{L}^B$ (Eqn.~\ref{eq:loss_binaural} in Sec.~\ref{approach:training} of main) and $\mathcal{L}^M$ (Eqn.~\ref{eq:loss_monaural} in Sec.~\ref{approach:training} of main) by using Adam~\citep{kingma2014adam} and a learning rate of $5e^{-4}$ until convergence.

To train the policies of Move2Hear, AV Navigator~\citep{chen2020soundspaces}, and Novelty~\citep{bellemare2016unifying} using PPO~\citep{schulman2017proximal} (`Training the Active Audio-Visual Controller' of Sec.~\ref{approach:training} in main), we weight the action loss by 1.0 and the value loss by 0.5. For $\pi^\mathcal{Q}$, we use an entropy loss on the policy distribution with a coefficient of 0.01 while for all other policies, we set the coefficient to 0.2. We train all policies with Adam~\citep{kingma2014adam} and a learning rate of $1e^{-4}$ for a total of 38 million policy prediction steps.